\pdfoutput=1

\documentclass[11pt]{article}

\usepackage{acl}

\usepackage{times}
\usepackage{latexsym}

\usepackage[T1]{fontenc}

\usepackage[utf8]{inputenc}

\usepackage{microtype}

\usepackage[pdftex]{graphicx}
\usepackage{subfigure}
\usepackage{float}
\graphicspath{./pics/}
\usepackage{multirow}
\usepackage{amsmath}
\usepackage{amsfonts}       
\usepackage{booktabs}
\newtheorem{challenge}{Challenge}

\title{Document-level Relation Extraction with Relation Correlations}

\author{Ridong Han$^{1,2,4,5}$ \ Tao Peng$^{1,2}$ \ Benyou Wang$^{4,5*}$ \ Lu Liu$^{2,3}$\thanks{\ \ Corresponding authors.} \ Xiang Wan$^{4,5}$ \\
  $^1$College of Computer Science and Technology, Jilin University \\
  $^2$Key Laboratory of Symbolic Computation and Knowledge Engineering \\
  of Ministry of Education, China \\
  $^3$College of Software, Jilin University \\
  $^4$Shenzhen Research Institute of Big Data, The Chinese University of Hong Kong, Shenzhen \\
  $^5$School of Data Science,  The Chinese University of Hong Kong, Shenzhen\\
  \texttt{hanrd20@mails.jlu.edu.cn} \\
}

\begin{document}
\maketitle
\begin{abstract}
Document-level relation extraction faces two overlooked challenges: long-tail problem and multi-label problem. 
Previous work focuses mainly on obtaining better contextual representations for entity pairs, hardly address the above challenges.
In this paper, we analyze the co-occurrence correlation of relations, and introduce it into DocRE task for the first time. 
We argue that the correlations can not only transfer knowledge between data-rich relations and data-scarce ones to assist in the training of tailed relations, but also reflect semantic distance guiding the classifier to identify semantically close relations for multi-label entity pairs. 
Specifically,
we use relation embedding as a medium, and propose two co-occurrence prediction sub-tasks from both coarse- and fine-grained perspectives
to capture relation correlations. Finally, the learned correlation-aware embeddings are used to guide the extraction of relational facts.
Substantial experiments on two popular DocRE datasets are conducted, and our method achieves superior results compared to baselines. Insightful analysis also demonstrates the potential of relation correlations to address the above challenges.
\end{abstract}

\section{Introduction}
Relation Extraction (RE) aims to recognize entity pairs of interest from a given text and classify their relationships. Earlier work mainly focuses on predicting the relationship of \textit{a given entity pair} within \textit{a single sentence}, i.e., sentence-level relation extraction (SentRE) \cite{zeng2014relation, lin2016neural, shikhar2018reside, han2022distantly}. 
This setting is too simplistic and is far from the real-world scenario, due to the fact that large amounts of relational facts are expressed in \textit{multiple sentences} (instead of a single sentence). For instance, more than 40.7\% of relational facts on DocRED (a popular relation extraction dataset) can only be extracted across multiple sentences \cite{yao2019docred}.

Compared to \textit{sentence}-level counterpart, \textit{document}-level relation extraction (DocRE) is more challenging, which identifies the relationships of all entity pairs within a document at once. 
Apart from the intrinsic characteristic, there are two other overlooked challenges, see \textbf{Challenge}~\ref{challenge:ltp} and ~\ref{challenge:mlp}.

\begin{challenge}
\textbf{Long-tail problem}:
\label{challenge:ltp}
the number of training triplets for different relations follows a long-tailed distribution, 
vanilla training on long-tailed relations will cause the model underfitted on tailed relations, yielding poor performance.
\end{challenge}

\begin{challenge}
\textbf{Multi-label problem}:
\label{challenge:mlp}
some entity pairs express multiple relations simultaneously, i.e., these co-expressed relations have partial semantic overlap, requiring more complex classification boundaries among them.
\end{challenge}

\begin{figure}[t]
	\centering
	\includegraphics[width=\linewidth]{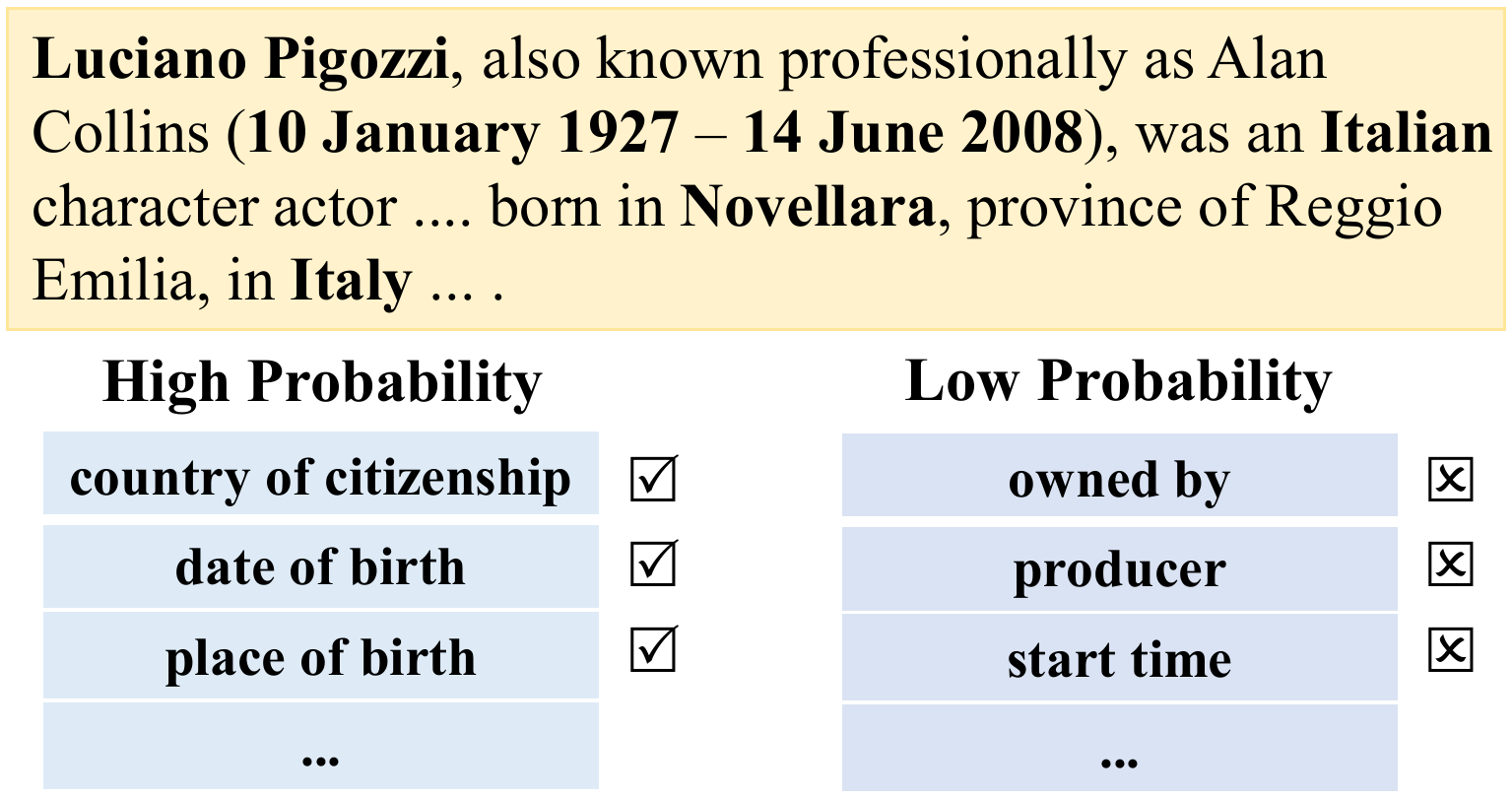}
	\caption{The given document is a biography. Relations ``\emph{date of birth}'' and ``\emph{place of birth}'' are more likely to occur than other relations such as ```\emph{producer}''.}
	\label{doc_example}
\end{figure}

For example, on DocRED dataset \cite{yao2019docred}, about 60 of the 96 non-\texttt{NA} relations are long-tailed, with less than 200 training triplets. F1 value of these data-scarce relations are even less than 1/3 of data-rich relations. Besides,
for entity pairs that express relations, at least 7\% have multiple relation labels. With the increase of labels, the F1 value decreases dramatically, by more than 50\%. These analyses indicate that the two above challenges cannot be ignored, see Section~\ref{data_exploration} for details.
However, existing studies rely heavily on Transformer \cite{zhou2021document, xie2022eider} or rule-based document graph \cite{nan2020reasoning, zeng2020double, xu2021discriminative, peng2022document} to obtain contextual representations, ignore 
the interdependencies among relations and therefore hardly deals with the above challenges.

\label{co-occurrence} 
In this paper, we introduce relation correlations into DocRE task.
We argue \textbf{relations in the real world barely appear independently, they usually interdepend on each other and appear together}. 
For example,  the given document in Figure~\ref{doc_example} is a personal biography about \emph{Luciano Pigozzi}, 
relations ``\emph{date of birth}'' and ``\emph{place of birth}'' are more likely to be expressed by the document than ``\emph{producer}''. Further, 
relation ``\emph{date of birth}'' is more likely to co-occur with ``\emph{place of birth}'' than ``\emph{producer}''. 
We refer to this co-occurrence between relations as \textit{Relation Correlations}, which are widely observed, as shown in Figure~\ref{correl_ppmi}.


Relation correlations can address the above challenges. 
On the one hand, 
for long-tailed relations, \textbf{their correlated relations may be data-rich}. By the correlations, data-rich relations can 
\textbf{transfer knowledge} to data-scarce ones, thus \textbf{assisting in the training of long-tail relations}.
On the other hand, 
for multi-label entity pairs, the same context expresses multiple relations, which means these co-expressed relations are semantically close. 
\textbf{The correlations can reflect the semantic distance among relations}, which can guide the classifier to \textbf{identify semantically close relations for multi-label pairs} and facilitate the delineation of decision boundaries.
See Section~\ref{data_exploration} for detailed analysis.

We first capture the correlations using relation embeddings, and then use the correlation-aware embeddings to guide the classifier to extract relational facts.
To improve the perception of relation correlations, we propose two auxiliary co-occurrence prediction sub-tasks from two perspectives, since co-occurrence could reflect correlations explicitly.
From the \textit{coarse-grained view}, relations as labels, Coarse-grained Relation Co-occurrence Prediction (CRCP) task judges whether one relation co-occurs with a set of other relations, based on the \textbf{relations} alone.
From the \textit{fine-grained view}, since relations depend on entity pairs to appear within a document, Fined-grained Relation Co-occurrence Prediction (FRCP) task determines whether one relation co-occurs with a set of other relations, based on \textbf{both relations and entity pairs}.

The learned relation embeddings which could perceive relation correlations thanks to the two auxiliary sub-tasks, accompanied by jointly-trained document contextual representations, are used to extract relational facts.

Substantial experiments on two popular datasets shows that the proposed method not only outperforms the existing baselines but also relieve the two challenges. Our model improves F1 by up to 6.49\% on long-tail relations, and up to 12.39\% on multi-label entity pairs, demonstrating the effectiveness of introducing relation correlations. We argue that this is the first work to introduce relation correlations into document-level relation extraction task, to the best of our knowledge.







\section{Methodology: Introduce Relation Correlations in DocRE}

\subsection{Motivations}  
\label{data_exploration}

To better illustrate the motivation, we first conduct some data exploration using statistical and experimental analysis.

\paragraph{Statistical Analysis}
We conduct statistical analysis on DocRED dataset \cite{yao2019docred}. 
The long-tailed distribution is particularly noticeable. More than half of the training triplets express only 6 of the 96 non-$\texttt{NA}$ relation types, and about 60 relation types has less than 200 training instances. In addition, for entity pairs that express relation(s), at least 7\% of them have multiple relation labels, which is extremely difficult to identify accurately.

\paragraph{Experimental Analysis}
We conduct experiments on long-tailed relations and multi-label entity pairs of DocRED, respectively, using our base model (\textbf{BERT}$_{\texttt{BASE}}$). See Section~\ref{base_model} and ~\ref{exp_settings} for details of model and experimental settings.

\begin{figure}[tbp]
	\centering
	\begin{minipage}{0.49\linewidth}
		\centering
		\includegraphics[width=\linewidth]{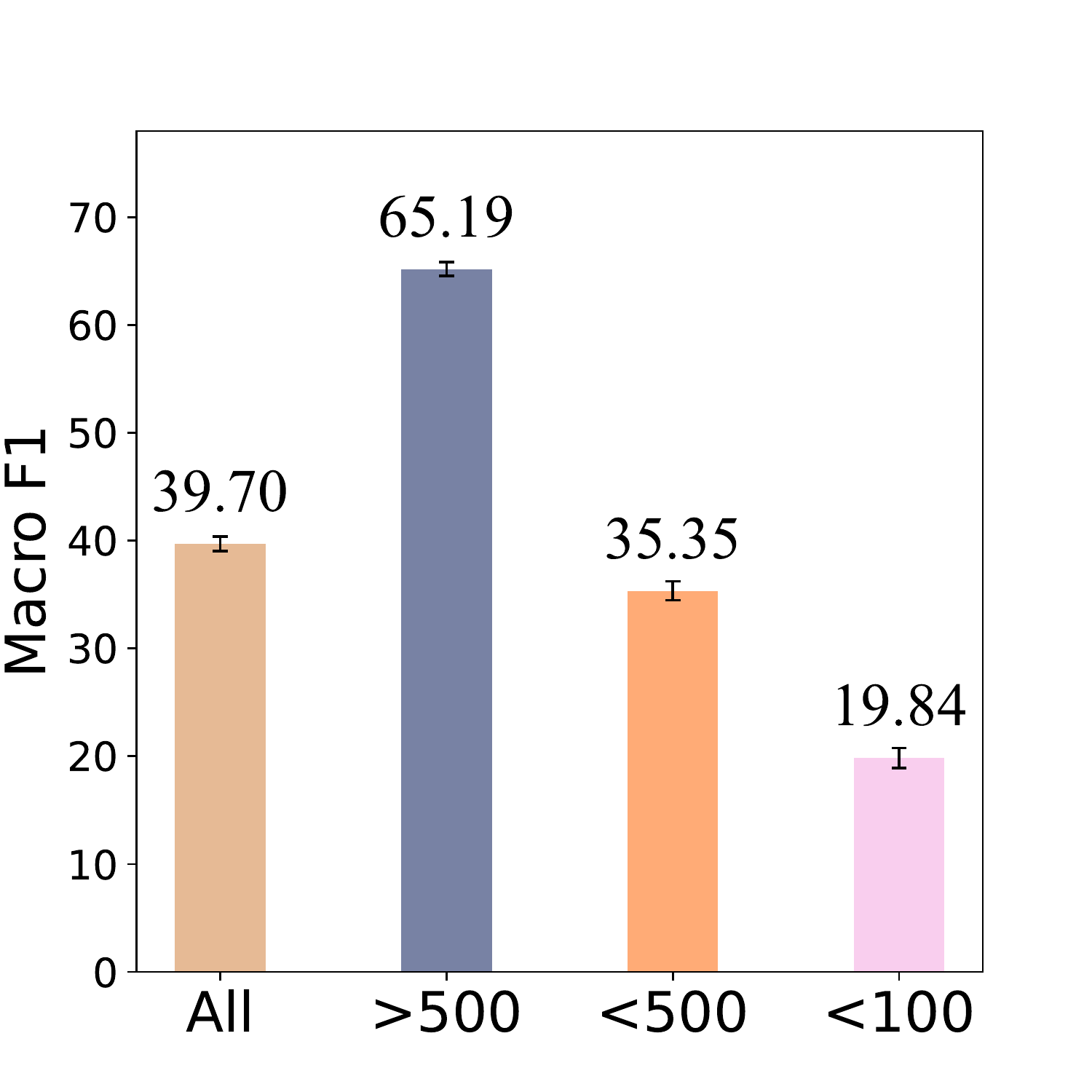}
		\label{inst_long}
	\end{minipage}
	\begin{minipage}{0.49\linewidth}
		\centering
		\includegraphics[width=\linewidth]{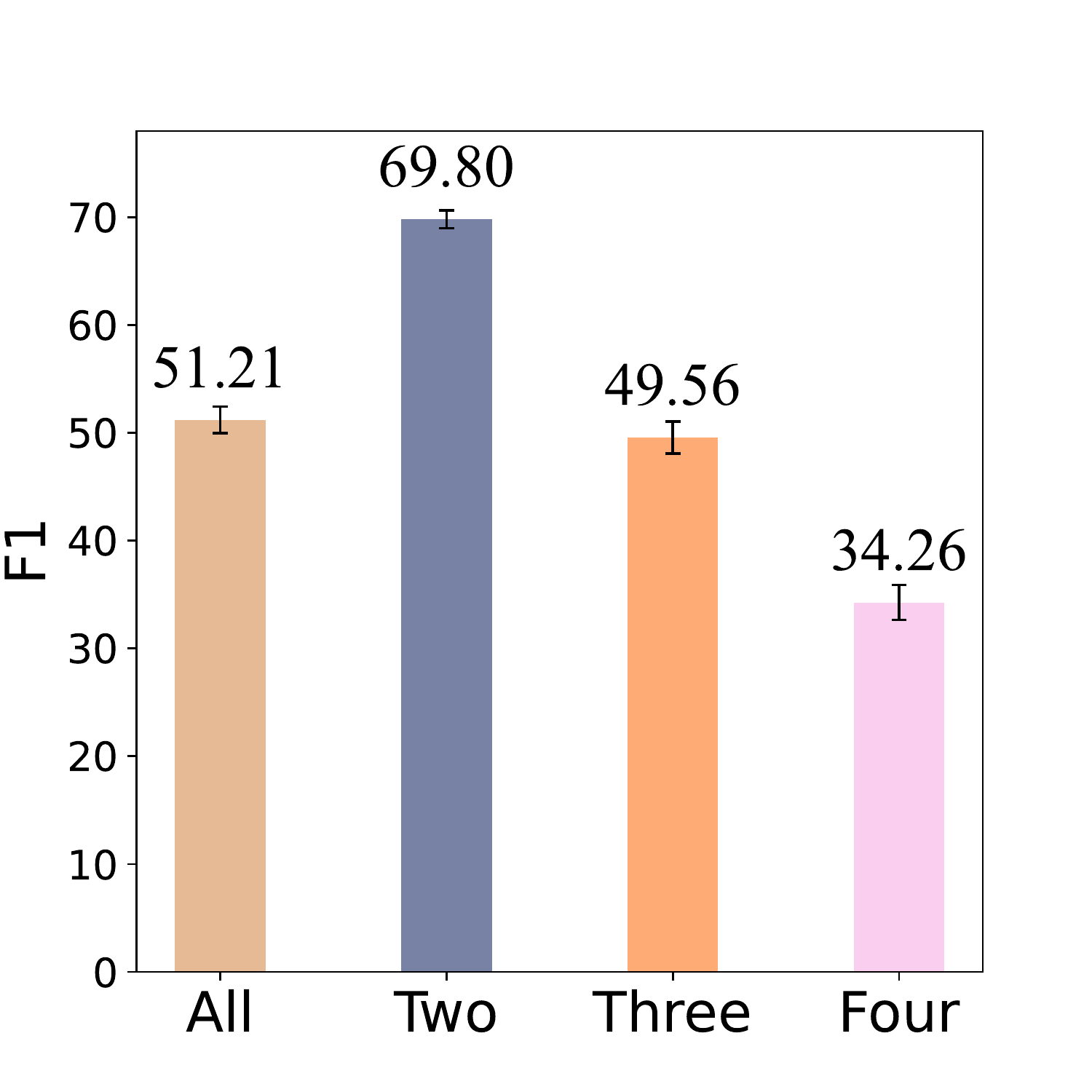}
		\label{inst_multi}
	\end{minipage}
    \caption{Left figure shows the results of \textbf{BERT}$_{\texttt{BASE}}$ on long-tailed relations; right figure is the results on multi-label entity pairs. We run 5 times with different seeds, and show the mean and standard deviation bar.}
    \label{de_bar_figures}
\end{figure}

For long-tailed relations, we calculate the macro-averaged F1 values for all relations (All), data-rich relations (>500) and long-tailed relations (<200 and <100). As shown in the left subfigure of Figure~\ref{de_bar_figures}, with the reduction of training instances, the performance decreases dramatically. The F1 of ``<100'' is even less than 1/3 of ``>500'' (19.84 $\texttt{vs.}$ 65.19). The poor performance of long-tailed relations severely limits the overall performance.

For multi-label entity pairs, we calculate the F1 values when the number of labels is 2, 3, 4, respectively, and report the overall macro-averaged F1. From the right subfigure of Figure~\ref{de_bar_figures}, we find that the higher the number of labels, the worse the performance. When there are more than 3 labels, the F1 value is substantially lower than the overall value (34.26 $\texttt{vs.}$51.21). 

\paragraph{Motivations}
Based on the above analyses, we obtain two observations: \textbf{long-tailed relations are under-trained} (i.e., \textbf{Challenge}~\ref{challenge:ltp}), and \textbf{the classification of multi-label entity pairs is more challenging than single-label pairs} (i.e., \textbf{Challenge}~\ref{challenge:mlp}).

\begin{figure}[t]
	\centering
    \includegraphics[width=\linewidth]{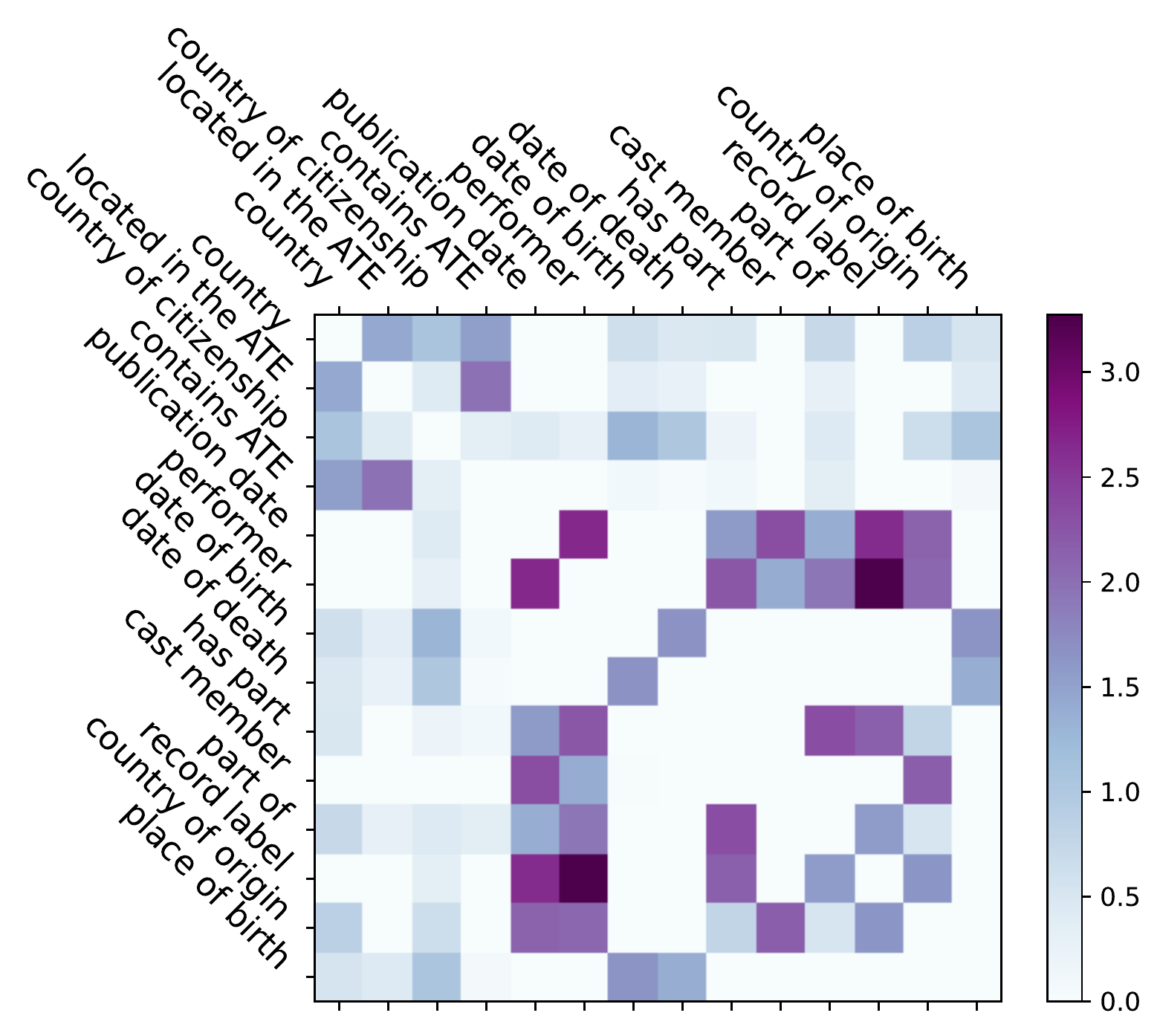}
	\caption{Relation correlations on DocRED. For better visual effect, we only show part of relations. See Appendix~\ref{appendix_a} for the complete figure.}
	\label{correl_ppmi}
\end{figure}

Existing work hardly deals with the above challenges. Therefore, we introduce relation correlations into DocRE task to alleviate or address them.
We observe and analyze the simplest co-occurrence correlation between relations. For example, in Figure~\ref{doc_example}, relation ``\textit{place of birth}'' is more likely to co-occur with ``\textit{place of date}'' than ``\textit{producer}''.
Further, 
we count the frequency of co-occurrence between every two relations on DocRED \cite{yao2019docred}, and use the positive point-wise mutual information values to measure the correlation of relations. As seen in Figure~\ref{correl_ppmi}, 
co-occurrence correlation between relations are widely observed. 

Relation correlations can alleviate the above challenges:
1) For long-tailed relations, they may be highly correlated with some data-rich relations. During training, we can transfer knowledge between long-tailed relations and their correlated relations, using data-rich head relations in the long-tail distribution to assist in the training of data-scarce tailed relations, 
thus alleviating the undertraining caused by insufficient training triplets. 
2) For multi-label entity pairs, the same context expresses multiple relations, which means these co-expressed relations have partial semantic overlap. The correlations can reflect the semantic distance among relations. If two relations co-occur frequently, then they are semantically closer than other relations. Such distance information can guide 
the classifier to identify semantically close relations for multi-label entity pairs, narrowing the search space and
facilitating the delineation of decision boundaries.

\subsection{DocRE with Relation Correlations}
\label{docre_with_rc}
\paragraph{Task Formulation} Suppose a document $d$ consists of $n$ tokens $\{w_i\}_{i=1}^n$, involving $p$ entities $\{e_i\}_{i=1}^p$. Each entity $e_i$ has $Q_i$ mentions $\{m_j\}_{j=1}^{Q_i}$. 
Document-level relation extraction task aims to predict one or more relation labels from $\mathcal{R}\cup\{\texttt{NA}\}$ for each entity pair $(e_s, e_o)_{s,o=1...p; s \neq o}$, where $\mathcal{R}$ is the set of pre-defined relations, and $\texttt{NA}$ denotes that no relation exists between two entities.

\begin{figure*}[t]
	\centering
    \includegraphics[width=0.95\linewidth]{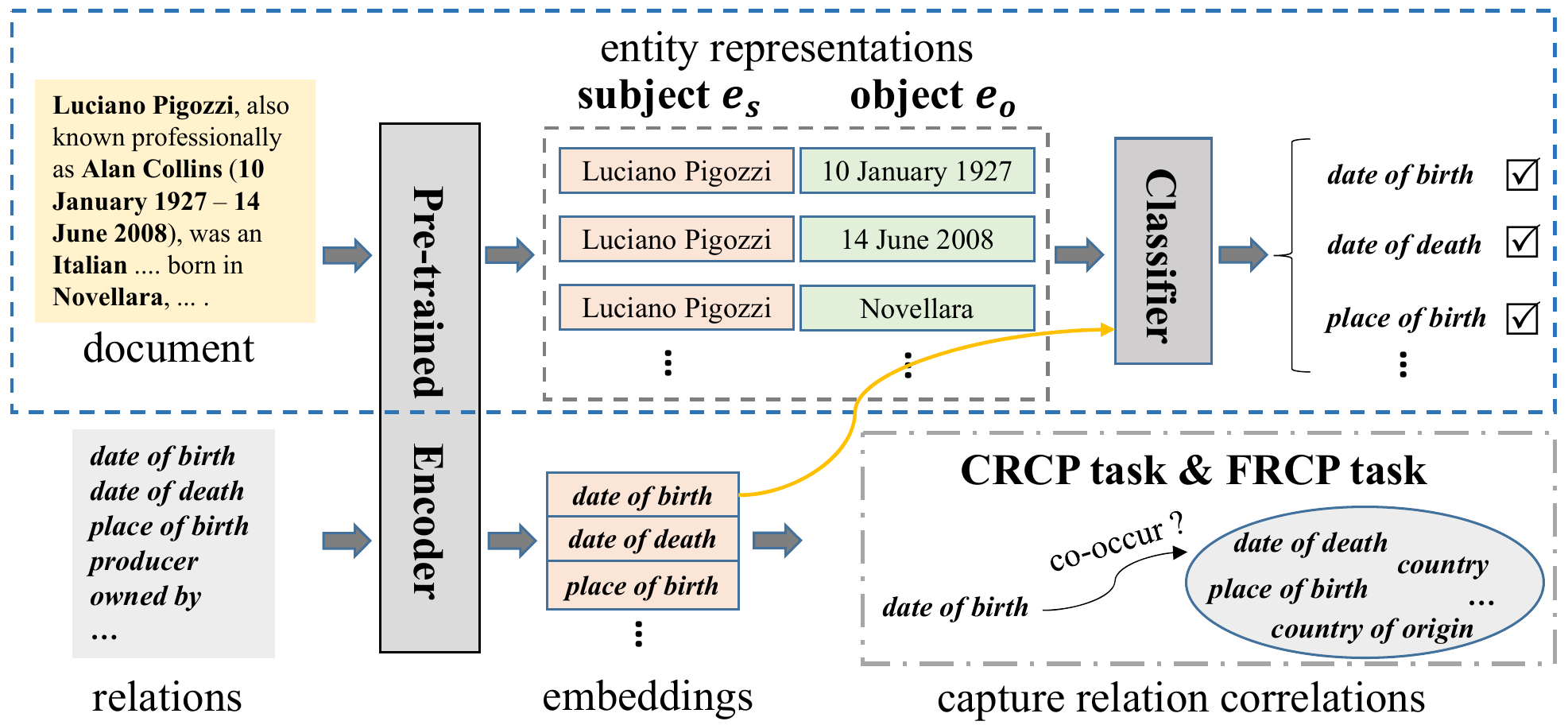}
	\caption{The overall framework of our method. The part circled by blue dashed line is the base model. The orange line indicates that relation embeddings learned by CRCP and FRCP tasks are used to guide relation classification.}
	\label{model_framework}
\end{figure*}

\paragraph{Overview} 
The overall framework of our model is shown in Figure~\ref{model_framework}.
We first jointly learn document contextual representations and relation embeddings through shared encoder (e.g., BERT), called joint embedding learning (JE).
Then, based on relation embeddings, we propose two auxiliary co-occurrence prediction tasks to capture relation correlations from both \textit{coarse}- and \textit{fine}-grained perspectives, including Coarse-grained Relation Co-occurrence Prediction (CRCP) and Fine-grained Relation Co-occurrence Prediction (FRCP).
Finally, the learned correlation-aware relation embeddings are used to enhance the feature representations of entity pairs, thus guiding relation classification. 

\paragraph{Joint Embedding Learning (JE)}
To obtain relation embeddings through shared encoder,
we combine the document with the sequence of all relations as the inputs to PLMs (e.g., BERT), i.e., ``$\texttt{[CLS]} w_1, w_2, ..., w_n \texttt{[SEP]} r_1, r_2, ...r_{|\mathcal{R}|} \texttt{[SEP]}$ '', where $r_i \in \mathcal{R}, i = 1, 2, ..., |\mathcal{R}|$. 
This method has two strengths: 1) Features and labels can be mapped into the same hidden space, eliminating the semantic gap between relation embeddings and document representations, and 2) Dependencies between document and labels can be automatically captured by stacked self-attention layers.

Since we only share the encoder of base model to obtain the initial relation embeddings, therefore our approach to capturing relation correlations is model-agnostic and can be built on top of existing models.
Next, we introduce our proposed sub-tasks in detail. 
Other implementation details are described in Section~\ref{implementation}.

\subsubsection{Coarse-grained Relation Co-occurrence Prediction (CRCP)}
From the coarse-grained perspective, based on \textbf{relation labels} alone,
we judge whether a relation co-occurs with a set of other relations.

For example, suppose the biography in Figure~\ref{doc_example} involves five relations, including \textit{date of birth}, \textit{place of birth}, \textit{date of death}, \textit{country} and \textit{country of origin}, CRCP task determines whether any one of them should co-occur with the other four relations. It assumes that \textbf{when the other four relations appear, the remaining one should also be expressed with high probability}.

Specifically,
one document $d$ has the expressed relation set $R^+$ and a unexpressed relation set $R^-$. 
For each $r$ in $R^+$, the pair $(r, R^+ -\{r\})$ serves as a positive example, marked as \textit{IsCo-Occur}. 
The task predicts whether target relation $r$ appears based on condition $R^+ - \{r\}$. 
By replacing target relation $r$ with one relation $r^-$ in $R^-$, negative examples $\{ (r^-, R^+ - \{r\}) \}$ can be constructed, marked as \textit{NotCo-Occur}.
The average of the embeddings of $R^+ -\{r\}$ is concatenated with the embedding of target relation ($r$ or $r^-$) as input feature of a binary classifier, obtaining co-occurrence probabilities.
\begin{align}
    p_{ij} &= P(r_j|d,R^+ -\{r_i\}) \notag \\
    &= \sigma (\texttt{\textbf{Linear}}([r_j; \texttt{\textbf{Avg}}(R^+ -\{r_i\})]))
    \label{co-occur-prob}
\end{align}

We use the binary cross-entropy to calculate the loss value $\mathcal{L}_{coarse}$ for CRCP task, i.e.,
\begin{align}
    \mathcal{L}_{coarse} = - \sum_{i,j}[&q_{ij}\log{p_{ij}} + \notag \\
    &(1-q_{ij})\log{(1-p_{ij})}]
    \label{equation_bce}
\end{align}
where $q_{ij}$ is the ground-truth (1 means \textit{IsCo-Occur} and 0 denotes \textit{NotCo-Occur}).

\subsubsection{Fine-grained Relation Co-occurrence Prediction (FRCP)}

From the fine-grained perspective, we determine whether a relation co-occurs with a set of other relations based on both \textbf{relations} and \textbf{entity pairs}, since relations depend on entity pairs to appear. 

To make relation embeddings better perceive the relation correlations, we propose a surrogate embedding for each entity pair called Relation-Aggregated Embeddings (RAE), which is aggregated using relation embeddings.

\paragraph{Relation-aggregated embedding (RAE)} 
For each entity pair $(e_s,e_o)$,
we first calculate matching scores between the pair and all relations using the token-to-relation attention matrix from the last transformer layer of shared encoder, and then use the scores to aggregate relation embeddings. For each entity $e_i$, the attention of all its mentions is averaged to be its attention $A_{r,e_i} \in \mathbb{R}^{N_h \times |r|}$, $N_h$ is the number of attention heads.
\begin{align}
    r_{(s,o)} = H_r^\texttt{T} \cdot  \texttt{\textbf{Norm}}(\sum_{k=1}^{N_h} A_{r,s}^k \circ A_{r,o}^k)  \label{r_s_o} 
\end{align}
where ``$\circ$'' denotes hadamard product and $H_r$ is the matrix consisting of all relation embeddings. 
Since attention values are always positive, we use sum normalization, i.e, $   \texttt{\textbf{Norm}}(\vec{x})  = \vec{x} / \textrm{sum}(\vec{x}) $.

\paragraph{Positive/negative examples} The resulting $r_{(s,o)}$ can be regarded as the relation representation of entity pair $(e_s, e_o)$.
Based on whether the entity pair expresses a relation, we divide all relation representations $\{ r_{(s,o)} \} _{(s,o) \in d}$ into two groups, non-\texttt{NA} relation set ${R_{\texttt{non-na}}}$ and \texttt{NA} relation set ${R_{\texttt{na}}}$. 
For each $r_{(s,o)}$ in $R_{\texttt{non-na}}$, 
the pair $(r_{(s,o)}, R_{\texttt{non-na}} - \{r_{(s,o)}\})$ serves as a positive example, marked as \textit{IsCo-Occur}.
The task uses the relations $R_{\texttt{non-na}} - \{r_{(s,o)}\}$ of conditional entity pairs to predict whether the relation $r_{(s,o)}$ of target entity pair co-occurs. By replacing the target relation $r_{(s,o)}$ with one item $r_{\texttt{na}}$ in ${R_{\texttt{na}}}$, 
negative examples $\{(r_{\texttt{na}}, R_{\texttt{non-na}} - \{r_{(s,o)}\})\}$ can be constructed, marked as \textit{NotCo-Occur}.

A similar thought could be seen in CBOW \cite{mikolov2013cbow}: using context vectors to predict the target. 
The same implementations as in Eqs.~\ref{co-occur-prob} and ~\ref{equation_bce} are used to obtain the co-occurrence probabilities and the loss value $\mathcal{L}_{fine}$.

\section{Specific Implementations}
\label{implementation}

\subsection{Base Model}
\label{base_model}
We build our base model (\textbf{BERT}$_{\texttt{BASE}}$) on the existing baselines \cite{wang2019fine-tune, zhou2021document}.
The document $d = \{w_i\}_{i=1}^n$ are fed into a pre-trained foundation model (e.g., BERT) to obtain the contextual representations $H=[h_1, h_2,...,h_n] \in \mathbb{R}^{n \times d_h}$, where $d_h$ is the embedding dimension. 
\begin{align}
    \label{encoder}
    H, A &= \texttt{PLMs}([w_1, w_2, ..., w_n]) 
\end{align}

All entity mentions are wrapped with a special token ``*'' \cite{zhang2017position, soares2019matching}, and the embedding of ``*'' in front of each mention is regarded as its representation $h_{m_j^i}$. Next, the log-sum-exp pooling \cite{jia20149document} is employed to get entity embedding $h_{e_i}$,
\begin{align}
    h_{e_i} &= \log{\sum_{j=1}^{N_{e_i}} \exp{(h_{m_j^i})}}
\end{align}
where $N_{e_i}$ is the number of mentions for entity $e_i$.

As \citet{zhou2021document}, the token-to-token attention matrix $A$ of the last transformer layer is used to aggregate context information $c_{(s,o)} \in \mathbb{R}^{d_h}$
for each entity pair $(e_s, e_o)$. For each entity $e_i$, the attention of all its mentions is averaged to be its attention $A_{e_i} \in \mathbb{R}^{N_h \times n}$. 
\begin{align}
    c_{(s,o)} = H^\texttt{T} \cdot \textrm{norm} (\sum_{k=1}^{N_h} A_s^k \circ A_o^k)
\end{align}

Finally, entity embeddings $h_{e_s}$ and $h_{e_o}$ are fused with contextual representation $c_{(s,o)}$, respectively. The resulting representation is fed to the grouped bilinear classifier \cite{tang2020orthogonal}. 
\begin{align}
    &f_s = \tanh{(W_sh_{e_s} + W_1c_{(s,o)})} \label{global_e_s}\\
    &f_o = \tanh{(W_oh_{e_o} + W_2c_{(s,o)})} \label{global_e_o}\\
    &P(r|e_s, e_o)=\sigma(\sum_{i=1}^k {f_s^i}^\texttt{T}W_r^if_o^i + b_r)  
\end{align}
where $[f^1_s; f^2_s;...;f^k_s] = f_s$,  $[f^1_o; f^2_o;...;f^k_o] = f_o$, 
$k$ is the number of groups and $\sigma$ is the sigmoid activation function.

\subsection{Base Model with Relation Correlations}
To leverage the correlation-aware relation embeddings to guide the classification, we use the relation-aggregated embedding $r_{(s,o)}$ from Eq.~\ref{r_s_o} as a complementary feature of entity pair $(e_s,e_o)$, by modifying Eq.~\ref{global_e_s} and Eq.~\ref{global_e_o} as follows:
\begin{align}
    &f_s = \tanh{(W_sh_{e_s} + W_3[c_{(s,o)};r_{(s,o)}])} \\
    &f_o = \tanh{(W_oh_{e_o} + W_4[c_{(s,o)};r_{(s,o)}])}
\end{align}
where $\{W_3,W_4\}\in \mathbb{R}^{2d_h \times d_h}$.

\paragraph{Training objective}
The main objective for DocRE is defined to minimize a binary cross-entropy loss $\mathcal{L}_{re}$ as Eq.~\ref{equation_bce}.
The overall loss can be calculated by:
\begin{align}
    &\mathcal{L}_{col} = \alpha \cdot \mathcal{L}_{coarse} + (1-\alpha) \cdot \mathcal{L}_{fine} \\
    &\mathcal{L} = (1 + \beta^2) \cdot \frac{\mathcal{L}_{re} \cdot \mathcal{L}_{col}}{(\beta^2 \cdot \mathcal{L}_{col}) + \mathcal{L}_{re}}
\end{align}
where $\alpha$ and $\beta$ are trade-off coefficients.

\section{Experiments and Analysis}
\subsection{Experimental Settings}
\label{dataset_intro}
\paragraph{Datasets} To evaluate our proposed model, we conduct experiments on two popular DocRE datasets, i.e., DocRED\cite{yao2019docred} and DWIE\cite{zaporojets2021dwie}. For DWIE, we employ the same pre-processing method as \citet{ru2021learning}.
The statistics are shown in Table~\ref{dataset}. 
Note that we do not choose two other commonly used datasets, i.e., CDR\cite{li2016cdr} and GDA\cite{wu2019gda}, because they have only one non-$\texttt{NA}$ relationship and are too polarized to model relation correlations.
\begin{table}[t]
    \centering
    \resizebox{0.95\linewidth}{!}{
    \begin{tabular}{lcccccc}
		\toprule
		Dataset & \#Train & \#Dev & \#Test & \#Relations \\
		\midrule
		DocRED & 3053 & 998 & 1000 & 96 \\
		DWIE & 602 & 98 & 99 & 65 \\
		\bottomrule
    \end{tabular}
    }
    \caption{Statistics of DocRED and DWIE datasets.}
    \label{dataset}
\end{table}

\paragraph{Implementation Details} We use the popular HuggingFace's Transformers library \citep{wolf2019transformers} to implement our model. For PLMs, we choose the pre-trained BERT-base-cased \citep{devlin2019bert}. During training, our model is optimized with AdamW \cite{loshchilov2019decoupled} using learning rate $\gamma$. And the warmup technique is also applied. All hyper-parameters are determined on the development set, some of which are listed in Table~\ref{parameters}.
while during inference, a global threshold determined on the validation set is used to decide whether the relation $r$ exists between $e_s$ and $e_o$. We search it from the list $[0.1,0.2,...,0.9]$ and pick the one with the highest F1 on the development set. All models are trained with 1 Tesla V100 GPU. 

\begin{table}[ht]
    \centering
    \begin{tabular}{lcccccc}
		\toprule
		Dataset     & DocRED & DWIE \\
		\midrule
		Batch Size & 4 & 4  \\
		\#Epoch & 50 & 30  \\
		lr for PLMs & 5e-5 & 5e-5 \\
		lr for others & 1e-4 & 1e-4 \\
		ratio for warmup & 6\% & 6\% \\
		$\alpha$,$\beta$ & 0.7,0.5 & 0.5,0.5 \\
		\bottomrule
    \end{tabular}
    \caption{Hyper-parameters in training.}
    \label{parameters}
\end{table}

\begin{table*}[htbp]
    \centering
    \resizebox{\linewidth}{!}{
        \begin{tabular}{lcccccccc}
    		\toprule
    		Dataset  & \multicolumn{4}{c}{DocRED}  & \multicolumn{4}{c}{DWIE} \\
    		\cmidrule(r){2-5} \cmidrule(r){6-9}
    		\ & \multicolumn{2}{c}{Dev} & \multicolumn{2}{c}{Test} & \multicolumn{2}{c}{Dev} & \multicolumn{2}{c}{Test} \\
    		\cmidrule(r){2-3}  \cmidrule(r){4-5} \cmidrule(r){6-7}  \cmidrule(r){8-9} 
    		Models & Ign. F1 & F1 & Ign. F1 & F1 & Ign. F1 & F1 & Ign. F1 & F1 \\
    		\midrule
    		CNN  & 37.99 & 43.45 & 36.44 & 42.33 & 37.65 & 47.73 & 34.65 & 46.14\\
    		LSTM  & 44.41 & 50.66 & 43.60 & 50.12 & 40.86 & 51.77 & 40.81 & 52.60\\
    		BiLSTM  & 45.12 & 50.95 & 44.73 & 51.06 & 40.46 & 51.92 & 42.03 & 54.47\\
    		Context-Aware & 44.84 & 51.10 & 43.93 & 50.64 & 42.06 & 53.05 & 45.37 & 56.58\\
    		CorefBERT$_{\texttt{BASE}}$ & 55.32 & 57.51 & 54.54 & 56.96 & 57.18 & 61.42 & 61.71 & 66.59 \\
    		GAIN$_{\texttt{BASE}}$ & 59.14 & 61.22 & 59.00 & 61.24 & 58.63 & 62.55 & 62.37 & 67.57 \\
    		SSAN$_{\texttt{BASE}}$ & 57.04 & 59.19 & 56.06 & 58.41 & 58.62 & 64.49 & 62.58 & 69.39 \\
    		ATLOP$_{\texttt{BASE}}$ & 59.22 & 61.09 & 59.31 & 61.30 & 59.03 & 64.82 & 62.09 & 69.94 \\
    		ERA$_{\texttt{BASE}}$ & 59.30 & 61.30 & 58.71 & 60.97 & - & - & - & - \\
    		ERACL$_{\texttt{BASE}}$ & 59.72 & 61.80 & 59.08 & 61.36 & - & - & - & - \\
    		RSMAN$_{\texttt{SSAN-BASE}}$ & 57.22 & 59.25 & 57.02 & 59.29 & 60.02 & 65.88 & 63.42 & 70.95\\
    		\midrule
    		\textbf{BERT}$_{\texttt{BASE}}$ & 58.09$\pm$0.11 & 60.10$\pm$0.12 & 58.03 & 60.20 & 58.40$\pm$0.26 & 63.38$\pm$0.33 & 62.92$\pm$0.64 & 69.12$\pm$0.56 \\
    		\textbf{BERT-Correl}$_{\texttt{BASE}}$ & \textbf{59.39$^\dagger\pm$0.13} & \textbf{61.34$^\dagger\pm$0.11} & \textbf{59.12} & \textbf{61.32} & \textbf{61.10$^\dagger\pm$0.54} & \textbf{65.73$^\dagger\pm$0.69} & \textbf{65.64$^\dagger\pm$0.52} & \textbf{71.56$^\dagger\pm$0.28} \\
    		\ & $\uparrow$1.30 & $\uparrow$1.24 & $\uparrow$1.09 & $\uparrow$1.12 & $\uparrow$1.70 & $\uparrow$2.35 & $\uparrow$2.72 & $\uparrow$2.44 \\ 
    		\midrule
    		\; \textbf{- CRCP} & 59.03$\pm$0.09 & 60.97$\pm$0.05 & 58.89 & 61.03 & 59.40$\pm$0.43 & 64.38$\pm$0.74 & 64.87$\pm$1.29 & 70.90$\pm$0.90 \\
    		\; \textbf{- FRCP} & 59.11$\pm$0.12 & 61.10$\pm$0.09 & 59.05 & 61.12 & 59.52$\pm$0.28 & 64.00$\pm$0.26 & 64.33$\pm$0.54 & 69.99$\pm$0.52 \\
    		\; \textbf{- CRCP and FRCP} & 58.33$\pm$0.23 & 60.24$\pm$0.20 & 58.38 & 60.45 & 58.97$\pm$0.13 & 63.92$\pm$0.47 & 63.72$\pm$0.64 & 69.72$\pm$0.61 \\
    		\; \; \; \textbf{- JE} (\textbf{BERT}$_{\texttt{BASE}}$) & 58.09$\pm$0.11 & 60.10$\pm$0.12 & 58.03 & 60.20 & 58.40$\pm$0.26 & 63.38$\pm$0.33 & 62.92$\pm$0.64 & 69.12$\pm$0.56 \\  
    		\bottomrule
        \end{tabular}
    }
    \caption{Main results (\%) on DocRED and DWIE datasets. We run 5 times with different random seeds, and report the mean and standard deviation (excluding the results of DocRED's test set, as it must be obtained through CodaLab site). For DocRED, the results are from their original paper, while for DWIE, we use the results reported in \citet{yu2022relation}. The subscript ``$\texttt{BASE}$'' indicates using BERT-base as the backbone encoder. $^\dagger$ indicates that the improvement is significant according to a two-sided T-Test ($p < 0.05$).}
    \label{overall_performance}
\end{table*}

\paragraph{Metrics} For overall performance, we use micro F1 and Ign. F1 in evaluation. The Ign. F1 denotes the F1 value after excluding entity pairs that appear in both train and development/test set. For long-tailed relations, we report the $\textbf{Macro@K}$ which means the macro-averaged F1 for relation types with frequency in training set less than $K$. Compared to micro-F1, macro-F1 treat all relation
types equally and can more accurately reflect the performance on long-tailed relations.
While for multi-label entity pairs, we report the micro-F1 value when there are two/three/four labels.

\paragraph{Baselines}
We compare our model with some typical or recent baselines, including CNN \cite{zeng2014relation}, LSTM/BiLSTM \cite{cai2016bidirectional}, Context-Aware \cite{sorokin2017context}, CorefBERT \cite{ye2020coref}, GAIN \cite{zeng2020double}, SSAN \cite{xu2021entity}, ATLOP \cite{zhou2021document}, ERA/ERACL \cite{du2022improving} and RSMAN \cite{yu2022relation}.
\label{exp_settings}
\subsection{Model Comparison Results}
\label{comparison}
Table~\ref{overall_performance} depicts the overall performance of our proposed model on two popular DocRE datasets, in comparison with baselines. We can find that our base model (\textbf{BERT}$_{\texttt{BASE}}$) outperforms some earlier BERT-based baselines, CorefBERT$_{\texttt{BASE}}$ \cite{ye2020coref} and SSAN \cite{xu2021entity}, on all result columns of DocRED and DWIE datasets. In other words, our base model yields strong performance compared to the earlier BERT-based models.
Further, our model \textbf{BERT-Correl}$_{\texttt{BASE}}$ improves the performance of \textbf{BERT}$_{\texttt{BASE}}$ by 1.30\%, 1.24\%, 1.09\%, 1.12\% for each column on DocRED dataset and 1.70\%, 2.35\%, 2.72\%, 2.44\% for each column on DWIE dataset. The improvement is consistent, which confirms the generalization of our approach.
All results pass the two-sided T-Test ($p < 0.05$), indicating that the improvement is significant.

\begin{table*}[ht]
    \centering
    \resizebox{\linewidth}{!}{
        \begin{tabular}{lcccccccc}
    		\toprule
    		Dataset  & \multicolumn{4}{c}{DocRED}  & \multicolumn{3}{c}{DWIE} \\
    		\cmidrule(r){2-5} \cmidrule(r){6-8}
    		\ & Macro & Macro@500 & Macro@200 & Macro@100 & Macro & Macro@100 & Macro@50 \\
    		\midrule
    		\textbf{BERT}$_{\texttt{BASE}}$ & 39.70$\pm$0.47 & 35.35$\pm$0.59 & 27.66$\pm$0.71 & 19.84$\pm$0.64 & 28.17$\pm$0.40 & 6.53$\pm$0.55 & 2.47$\pm$0.50 \\
    		\textbf{BERT-Correl}$_{\texttt{BASE}}$ & \textbf{40.81$^\dagger\pm$0.35} & \textbf{36.55$^\dagger\pm$0.40} & \textbf{28.76$^\dagger\pm$0.63} & \textbf{21.38$^\dagger\pm$0.96} & \textbf{32.80$^\dagger\pm$1.25} & \textbf{13.02$^\dagger\pm$1.46} & \textbf{8.59$^\dagger\pm$1.73} \\
    		\ & $\uparrow$1.11 & $\uparrow$1.20 & $\uparrow$1.10 & $\uparrow$1.54 & $\uparrow$4.63 & $\uparrow$6.49 & $\uparrow$6.12 \\
    		\midrule
    		\; \textbf{- CRCP} & 40.48$\pm$0.32 & 36.18$\pm$0.36 & 28.32$\pm$0.39 & 20.56$\pm$0.54 & 30.52$\pm$0.63 & 10.24$\pm$0.94 & 5.98$\pm$1.47 \\
    		\; \textbf{- FRCP} & 40.52$\pm$0.32 & 36.22$\pm$0.32 & 28.53$\pm$0.51 & 20.52$\pm$0.67 & 30.46$\pm$0.99 & 10.47$\pm$1.46 & 6.19$\pm$1.47  \\
    		\; \textbf{- CRCP and FRCP} & 39.77$\pm$0.57 & 35.47$\pm$0.62 & 27.63$\pm$0.64 & 19.93$\pm$0.52 & 27.81$\pm$0.41 & 5.62$\pm$0.43 & 1.88$\pm$0.64 \\
    		\; \; \; \textbf{- JE} (\textbf{BERT}$_{\texttt{BASE}}$) & 39.70$\pm$0.47 & 35.35$\pm$0.59 & 27.66$\pm$0.71 & 19.84$\pm$0.64 & 28.17$\pm$0.40 & 6.53$\pm$0.55 & 2.47$\pm$0.50 \\  
    		\bottomrule
        \end{tabular}
    }
    \caption{Evaluation on long-tailed relations. We run 5 times with different random seeds on the development set, and report the mean and standard deviation. Labels of DocRED's test set are not accessible, so the values on test set cannot be displayed. $^\dagger$ indicates that the improvement is significant according to a two-sided T-Test ($p < 0.05$).}
    \label{tail_performance}
\end{table*}

\begin{table*}[ht]
    \centering
    \resizebox{\linewidth}{!}{
        \begin{tabular}{lcccccccc}
    		\toprule
    		Dataset  & \multicolumn{4}{c}{DocRED}  & \multicolumn{3}{c}{DWIE} \\
    		\cmidrule(r){2-5} \cmidrule(r){6-8}
    		\ & Two & Three & Four & Macro F1 & Two & Three & Macro F1 \\
    		\midrule
    		\textbf{BERT}$_{\texttt{BASE}}$ & 69.80$\pm$0.69 & 49.56$\pm$1.49 & 34.26$\pm$1.62 & 51.21$\pm$1.17 & 70.13$\pm$0.59 & 77.11$\pm$1.03 & 73.62$\pm$0.38 \\
    		\textbf{BERT-Correl}$_{\texttt{BASE}}$ & \textbf{70.97$^\dagger\pm$0.73} & \textbf{55.16$^\dagger\pm$0.58} & \textbf{46.65$^\dagger\pm$1.95} & \textbf{57.59$^\dagger\pm$0.63} & \textbf{73.98$^\dagger\pm$1.63} & \textbf{79.29$^\dagger\pm$1.41} & \textbf{76.63$^\dagger\pm$0.86} \\
    		\ & $\uparrow$1.17 & $\uparrow$5.60 & $\uparrow$12.39 & $\uparrow$6.38 & $\uparrow$3.85 & $\uparrow$2.18 & $\uparrow$3.01 \\
    		\midrule
    		\; \textbf{- CRCP} & 70.53$\pm$1.31 & 56.40$\pm$1.22 & 39.99$\pm$0.82 & 55.64$\pm$0.90 & 71.91$\pm$1.20 & 78.13$\pm$1.34 & 75.02$\pm$1.12 \\
    		\; \textbf{- FRCP} & 70.64$\pm$0.83 & 55.42$\pm$1.81 & 41.88$\pm$1.33 & 55.98$\pm$0.57 & 71.17$\pm$0.57 & 78.30$\pm$1.33 & 74.74$\pm$0.71  \\
    		\; \textbf{- CRCP and FRCP} & 69.92$\pm$0.80 & 51.49$\pm$1.92 & 40.00$\pm$0.01 & 53.80$\pm$0.72 & 70.01$\pm$0.74 & 76.22$\pm$0.82 & 73.11$\pm$0.61 \\
    		\; \; \; \textbf{- JE} (\textbf{BERT}$_{\texttt{BASE}}$) & 69.80$\pm$0.83 & 49.56$\pm$1.49 & 34.26$\pm$1.62 & 51.21$\pm$1.23 & 70.13$\pm$0.59 & 77.11$\pm$1.03 & 73.62$\pm$0.38 \\  
    		\bottomrule
        \end{tabular}
    }
    \caption{Results on multi-label entity pairs. We run 5 times with different random seeds, and report the mean and standard deviation. $^\dagger$ indicates that the improvement is significant according to a two-sided T-Test ($p < 0.05$).}
    \label{multi_label_performance}
\end{table*}

\subsection{Ablation Study}
To show the efficacy of each component, we conduct ablation studies on both DocRED and DWIE, by turning off one component at a time. The results are shown at the bottom of Table~\ref{overall_performance}. We can find that without CRCP and FRCP, the model performance has no or little improvement over the base model \textbf{BERT-Correl}$_{\texttt{BASE}}$. (Please refer to the bottom two rows of Table~\ref{overall_performance}, here $\textbf{-JE}$ indicates the model for further disabling the JE module when CRCP and FRCP are not enabled, which is the same as \textbf{BERT-Correl}$_{\texttt{BASE}}$.) When either CRCP or FRCP is enabled, the model achieves a significant performance improvement over \textbf{BERT-Correl}$_{\texttt{BASE}}$. This observation also demonstrates that our proposed sub-tasks, CRCP and FRCP, are both very effective, and that introducing relation correlations into DocRE scenario is highly powerful. Moreover, when we enable both CRCP and FRCP, the performance is improved by at least 1.09\% on DocRED and at least 1.70\% on DWIE.

\subsection{Analysis and Visualization}
\paragraph{Results on Long-tailed Relations}
To explore the impact of relation correlations on long-tailed relations, we conduct experiments on the tailed relation types which has fewer than $K$ training instances/ triplets. Then we report the macro-averaged F1 ($\textbf{Macro@K}$) for relation types with less than $K$ training instances in Table~\ref{tail_performance}. $K$ selects values of 500, 200 and 100 for DocRED, while for DWIE, $K$ selects values of 100 and 50. 
``$\textbf{Macro}$'' denotes the macro-averaged F1 for all relation types.
We can find that our model \textbf{BERT-Correl}$_{\texttt{BASE}}$ achieves consistent performance gains on long-tail relations for both DocRED and DWIE datasets, mitigating the Long-tail problem. The fewer training instances, the greater the improvement.
In particular, for DWIE dataset, the performance improvement of macro-averaged F1 on tailed relations is over 6\%, which indicates relation correlations has great potential for addressing long-tail relation classification. 
The two-sided T-Test are also employed to verify the significance of the improvement.

\paragraph{Results on Multi-Label Entity Pairs}
We select all entity pairs expressing multiple relation types from the development set, and evaluate our model on them. 
Since the labels of DocRED's test set are not available, the results on test set cannot be reported, we just report the results on the development set. Detailed results are shown in Table~\ref{multi_label_performance}. 
Here multiple labels of an entity pair are individually judged to be correct or not, and counted separately. 
If an entity pair has 2 labels, then there are two relational triplets to evaluate.
We find that relation correlations delivers consistent improvement on multi-label entity pairs of two datasets. The higher the number of labels, the more significant the improvement. In particular, for entity pairs with 4 labels on DocRED, our \textbf{BERT-Correl}$_{\texttt{BASE}}$ improves by 12.39\%.

\begin{figure}[t]
	\centering
	\includegraphics[width=\linewidth]{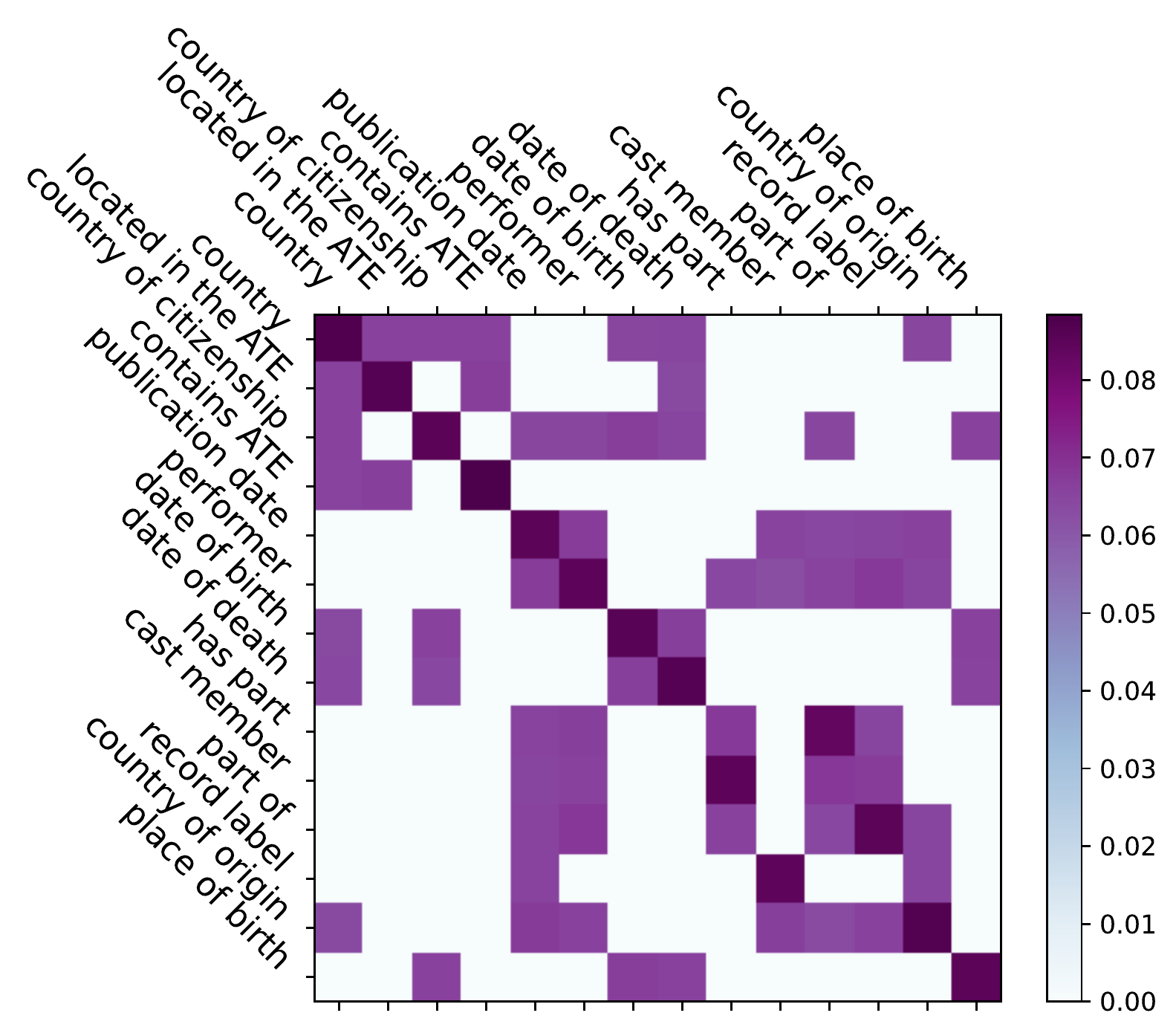}
	\caption{Relation correlations learned by our model on DocRED. For better visual effect, we only show part of relations as Figure~\ref{correl_ppmi}.}
	\label{correl_vis}
\end{figure}

\paragraph{Visualization of Relation Correlations}
We use the relation embeddings learned by our model on DocRED for visualization. We employ dot-product to calculate the relation similarity matrix, and visualize it in Figure~\ref{correl_vis}. For each row, we only display top-8 correlated relations for better visual effect. We find Figure~\ref{correl_vis} similar to Figure~\ref{correl_ppmi}, implying that we do capture large amounts of correlation knowledge based on CRCP task and FRCP task.

\section{Related Work}
\paragraph{Sentence-level Relation Extraction} 
Sentence-level relation extraction mainly includes 
CNN-based methods \cite{zeng2014relation, santos2015classifying, zeng2015distant, peng2022distantly}, RNN-based methods \cite{zhang2015bidirectional, cai2016bidirectional}, GNN-based methods \cite{zhang2018graph, zhu2019graph} and attention-based methods \cite{lin2016neural, dai2019feature, ye2019distant, yuan2019cross, han2022distantly}. These methods mainly benefit from local features \cite{peng2020learning} such as entity type, entity position, etc. In DocRE, these local features are far from sufficient, the global cross-sentence dependencies and reasoning are necessary.

\paragraph{Document-level Relation Extraction} 
Existing work on document-level relation extraction can be mainly divided into three categories, including sequence-based models \cite{yao2019docred}, graph-based models \cite{nan2020reasoning, zeng2020double, xu2021discriminative, xu2021document, li2022document, peng2022document} and Transformer-based models \cite{zhou2021document, yu2022relation, xie2022eider}. The sequence-based models use CNN \cite{googfellow2016deep} or LSTM \cite{schuster1997lstm} to encode the entire document directly, just copying the SentRE models and yielding poor performance. The graph-based models construct document graphs based on heuristic rules, encoding with graph neural networks \cite{petar2018gat}, achieving high performance. Unlike the graph-based models that rely heavily on handcrafted rules, the Transformer-based models leverage the capability for long-range dependencies of PLMs to directly obtain entity embeddings and contextual representations.

\paragraph{Relation Correlations} 
None of the above methods consider relation correlations. 
Similar to label correlations in the multi-label learning field \cite{zhu2018multi, zhang2014areview}, relation correlations are originally proposed by \citet{jin2020relation}, they define this phenomenon as ``Relation of Relations (ROR)'' through statistical analysis. Then \citet{peng2022distantly} and \citet{han2022distantly} introduce relation correlations into the hierarchical relation extraction task under the distant supervision paradigm, and achieve state-of-the-art performance. However, both studies perform sentence-level RE where the relation types have a hierarchical taxonomic structure. As clues, the structure can be directly used to model the correlations. In contrast, this paper focuses on DocRE task that the taxonomic structure does not exist, which is more challenging to capture the correlations.

\section{Conclusion}
In this paper, we analyze DocRE task in depth, and draw two observations: 1) The long-tailed relations are under-trained (i.e., challenge~\ref{challenge:ltp}), and 2) The classification of multi-label entity pair is more challenging than single-label pairs (i.e., challenge~\ref{challenge:mlp}). We analyze the co-occurrence correlation of relations, and introduce it into DocRE task.

We deem that relation correlations can alleviate the above challenges: 1) The correlations can transfer knowledge between data-rich relations and data-scarce long-tailed ones, assisting in the training of tailed relations, and 2) The correlations can reflect semantic distance among relations, and guide the classifier to identify semantically close relations for multi-label pairs.
Specifically, we first jointly learn the relation embeddings with document representations.
Then,
two auxiliary co-occurrence prediction sub-tasks (i.e., CRCP and FRCP) are designed to capture the correlations using relation embeddings. 
Finally, the learned correlation-aware relation embeddings are used to guide the classification. 
We conduct experiments on two popular datasets.
Relation correlations not only bring consistent improvements in overall RE performance, but also show strong potential for long-tail relations and multi-label entity pairs.



\bibliography{custom}

\begin{thebibliography}{48}
\expandafter\ifx\csname natexlab\endcsname\relax\def\natexlab#1{#1}\fi

\bibitem[{Cai et~al.(2016)Cai, Zhang, and Wang}]{cai2016bidirectional}
Rui Cai, Xiaodong Zhang, and Houfeng Wang. 2016.
\newblock \href {https://doi.org/10.18653/v1/p16-1072} {Bidirectional recurrent
  convolutional neural network for relation classification}.
\newblock In \emph{Proceedings of the 54th Annual Meeting of the Association
  for Computational Linguistics, {ACL}}. The Association for Computer
  Linguistics.

\bibitem[{Dai et~al.(2019)Dai, Xu, and Song}]{dai2019feature}
Longqi Dai, Bo~Xu, and Hui Song. 2019.
\newblock \href {https://doi.org/10.1007/978-3-030-32233-5\_15} {Feature-level
  attention based sentence encoding for neural relation extraction}.
\newblock In \emph{Natural Language Processing and Chinese Computing - 8th
  {CCF} International Conference, {NLPCC}}, volume 11838 of \emph{Lecture Notes
  in Computer Science}, pages 184--196. Springer.

\bibitem[{Devlin et~al.(2019)Devlin, Chang, Lee, and
  Toutanova}]{devlin2019bert}
Jacob Devlin, Ming{-}Wei Chang, Kenton Lee, and Kristina Toutanova. 2019.
\newblock \href {https://doi.org/10.18653/v1/n19-1423} {{BERT:} pre-training of
  deep bidirectional transformers for language understanding}.
\newblock In \emph{Proceedings of the 2019 Conference of the North American
  Chapter of the Association for Computational Linguistics: Human Language
  Technologies, {NAACL-HLT}}, pages 4171--4186. Association for Computational
  Linguistics.

\bibitem[{dos Santos et~al.(2015)dos Santos, Xiang, and
  Zhou}]{santos2015classifying}
C{\'{\i}}cero~Nogueira dos Santos, Bing Xiang, and Bowen Zhou. 2015.
\newblock \href {https://doi.org/10.3115/v1/p15-1061} {Classifying relations by
  ranking with convolutional neural networks}.
\newblock In \emph{Proceedings of the 53rd Annual Meeting of the Association
  for Computational Linguistics and the 7th International Joint Conference on
  Natural Language Processing of the Asian Federation of Natural Language
  Processing, {ACL}}, pages 626--634. The Association for Computer Linguistics.

\bibitem[{Du et~al.(2022)Du, Ma, Wu, Wu, Zhang, Long, and Ji}]{du2022improving}
Yangkai Du, Tengfei Ma, Lingfei Wu, Yiming Wu, Xuhong Zhang, Bo~Long, and
  Shouling Ji. 2022.
\newblock \href {https://doi.org/10.48550/arXiv.2205.10511} {Improving long
  tailed document-level relation extraction via easy relation augmentation and
  contrastive learning}.
\newblock \emph{CoRR}, abs/2205.10511.

\bibitem[{Goodfellow et~al.(2016)Goodfellow, Bengio, and
  Courville}]{googfellow2016deep}
Ian~J. Goodfellow, Yoshua Bengio, and Aaron~C. Courville. 2016.
\newblock \href {http://www.deeplearningbook.org/} {\emph{Deep Learning}}.
\newblock Adaptive Computation and Machine Learning. {MIT} Press.

\bibitem[{Han et~al.(2022)Han, Peng, Han, Cui, and Liu}]{han2022distantly}
Ridong Han, Tao Peng, Jiayu Han, Hai Cui, and Lu~Liu. 2022.
\newblock \href {https://doi.org/10.1016/j.neunet.2022.04.019} {Distantly
  supervised relation extraction via recursive hierarchy-interactive attention
  and entity-order perception}.
\newblock \emph{Neural Networks}, 152:191--200.

\bibitem[{Jia et~al.(2019)Jia, Wong, and Poon}]{jia20149document}
Robin Jia, Cliff Wong, and Hoifung Poon. 2019.
\newblock \href {https://doi.org/10.18653/v1/n19-1370} {Document-level n-ary
  relation extraction with multiscale representation learning}.
\newblock In \emph{Proceedings of the 2019 Conference of the North American
  Chapter of the Association for Computational Linguistics: Human Language
  Technologies, {NAACL-HLT}}, pages 3693--3704. Association for Computational
  Linguistics.

\bibitem[{Jin et~al.(2020)Jin, Yang, Qiu, and Zhang}]{jin2020relation}
Zhijing Jin, Yongyi Yang, Xipeng Qiu, and Zheng Zhang. 2020.
\newblock \href {http://arxiv.org/abs/2006.03719} {Relation of the relations:
  {A} new paradigm of the relation extraction problem}.
\newblock \emph{CoRR}, abs/2006.03719.

\bibitem[{Li et~al.(2016)Li, Sun, Johnson, Sciaky, Wei, Leaman, Davis,
  Mattingly, Wiegers, and Lu}]{li2016cdr}
Jiao Li, Yueping Sun, Robin~J. Johnson, Daniela Sciaky, Chih{-}Hsuan Wei,
  Robert Leaman, Allan~Peter Davis, Carolyn~J. Mattingly, Thomas~C. Wiegers,
  and Zhiyong Lu. 2016.
\newblock \href {https://doi.org/10.1093/database/baw068} {Biocreative {V}
  {CDR} task corpus: a resource for chemical disease relation extraction}.
\newblock \emph{Database J. Biol. Databases Curation}, 2016.

\bibitem[{Li et~al.(2022)Li, Lian, Lu, and Tang}]{li2022document}
Lishuang Li, Ruiyuan Lian, Hongbin Lu, and Jingyao Tang. 2022.
\newblock \href {https://aclanthology.org/2022.coling-1.183} {Document-level
  biomedical relation extraction based on multi-dimensional fusion information
  and multi-granularity logical reasoning}.
\newblock In \emph{Proceedings of the 29th International Conference on
  Computational Linguistics, {COLING}}, pages 2098--2107. International
  Committee on Computational Linguistics.

\bibitem[{Lin et~al.(2016)Lin, Shen, Liu, Luan, and Sun}]{lin2016neural}
Yankai Lin, Shiqi Shen, Zhiyuan Liu, Huanbo Luan, and Maosong Sun. 2016.
\newblock \href {https://doi.org/10.18653/v1/p16-1200} {Neural relation
  extraction with selective attention over instances}.
\newblock In \emph{Proceedings of the 54th Annual Meeting of the Association
  for Computational Linguistics, {ACL}}. The Association for Computer
  Linguistics.

\bibitem[{Loshchilov and Hutter(2019)}]{loshchilov2019decoupled}
Ilya Loshchilov and Frank Hutter. 2019.
\newblock \href {https://openreview.net/forum?id=Bkg6RiCqY7} {Decoupled weight
  decay regularization}.
\newblock In \emph{Proceedings of the 7th International Conference on Learning
  Representations, {ICLR}}. OpenReview.net.

\bibitem[{Mikolov et~al.(2013)Mikolov, Chen, Corrado, and
  Dean}]{mikolov2013cbow}
Tom{\'{a}}s Mikolov, Kai Chen, Greg Corrado, and Jeffrey Dean. 2013.
\newblock \href {http://arxiv.org/abs/1301.3781} {Efficient estimation of word
  representations in vector space}.
\newblock In \emph{1st International Conference on Learning Representations,
  {ICLR}, Workshop Track Proceedings}.

\bibitem[{Nan et~al.(2020)Nan, Guo, Sekulic, and Lu}]{nan2020reasoning}
Guoshun Nan, Zhijiang Guo, Ivan Sekulic, and Wei Lu. 2020.
\newblock \href {https://doi.org/10.18653/v1/2020.acl-main.141} {Reasoning with
  latent structure refinement for document-level relation extraction}.
\newblock In \emph{Proceedings of the 58th Annual Meeting of the Association
  for Computational Linguistics, {ACL}}, pages 1546--1557. Association for
  Computational Linguistics.

\bibitem[{Peng et~al.(2020)Peng, Gao, Han, Lin, Li, Liu, Sun, and
  Zhou}]{peng2020learning}
Hao Peng, Tianyu Gao, Xu~Han, Yankai Lin, Peng Li, Zhiyuan Liu, Maosong Sun,
  and Jie Zhou. 2020.
\newblock \href {https://doi.org/10.18653/v1/2020.emnlp-main.298} {Learning
  from context or names? an empirical study on neural relation extraction}.
\newblock In \emph{Proceedings of the 2020 Conference on Empirical Methods in
  Natural Language Processing, {EMNLP}}, pages 3661--3672. Association for
  Computational Linguistics.

\bibitem[{Peng et~al.(2022{\natexlab{a}})Peng, Han, Cui, Yue, Han, and
  Liu}]{peng2022distantly}
Tao Peng, Ridong Han, Hai Cui, Lin Yue, Jiayu Han, and Lu~Liu.
  2022{\natexlab{a}}.
\newblock \href {https://doi.org/10.1016/j.knosys.2021.107637} {Distantly
  supervised relation extraction using global hierarchy embeddings and local
  probability constraints}.
\newblock \emph{Knowledge-Based Systems.}, 235:107637.

\bibitem[{Peng et~al.(2022{\natexlab{b}})Peng, Zhang, and
  Xu}]{peng2022document}
Xingyu Peng, Chong Zhang, and Ke~Xu. 2022{\natexlab{b}}.
\newblock \href {https://doi.org/10.24963/ijcai.2022/601} {Document-level
  relation extraction via subgraph reasoning}.
\newblock In \emph{Proceedings of the Thirty-First International Joint
  Conference on Artificial Intelligence, {IJCAI}}, pages 4331--4337. ijcai.org.

\bibitem[{Ru et~al.(2021)Ru, Sun, Feng, Qiu, Zhou, Zhang, Yu, and
  Li}]{ru2021learning}
Dongyu Ru, Changzhi Sun, Jiangtao Feng, Lin Qiu, Hao Zhou, Weinan Zhang, Yong
  Yu, and Lei Li. 2021.
\newblock \href {https://doi.org/10.18653/v1/2021.emnlp-main.95} {Learning
  logic rules for document-level relation extraction}.
\newblock In \emph{Proceedings of the 2021 Conference on Empirical Methods in
  Natural Language Processing, {EMNLP}}, pages 1239--1250. Association for
  Computational Linguistics.

\bibitem[{Schuster and Paliwal(1997)}]{schuster1997lstm}
Mike Schuster and Kuldip~K. Paliwal. 1997.
\newblock \href {https://doi.org/10.1109/78.650093} {Bidirectional recurrent
  neural networks}.
\newblock \emph{{IEEE} Trans. Signal Process.}, 45(11):2673--2681.

\bibitem[{Soares et~al.(2019)Soares, FitzGerald, Ling, and
  Kwiatkowski}]{soares2019matching}
Livio~Baldini Soares, Nicholas FitzGerald, Jeffrey Ling, and Tom Kwiatkowski.
  2019.
\newblock \href {https://doi.org/10.18653/v1/p19-1279} {Matching the blanks:
  Distributional similarity for relation learning}.
\newblock In \emph{Proceedings of the 57th Conference of the Association for
  Computational Linguistics, {ACL}}, pages 2895--2905. Association for
  Computational Linguistics.

\bibitem[{Sorokin and Gurevych(2017)}]{sorokin2017context}
Daniil Sorokin and Iryna Gurevych. 2017.
\newblock \href {https://doi.org/10.18653/v1/d17-1188} {Context-aware
  representations for knowledge base relation extraction}.
\newblock In \emph{Proceedings of the 2017 Conference on Empirical Methods in
  Natural Language Processing, {EMNLP}}, pages 1784--1789. Association for
  Computational Linguistics.

\bibitem[{Tang et~al.(2020)Tang, Huang, Wang, He, and
  Zhou}]{tang2020orthogonal}
Yun Tang, Jing Huang, Guangtao Wang, Xiaodong He, and Bowen Zhou. 2020.
\newblock \href {https://doi.org/10.18653/v1/2020.acl-main.241} {Orthogonal
  relation transforms with graph context modeling for knowledge graph
  embedding}.
\newblock In \emph{Proceedings of the 58th Annual Meeting of the Association
  for Computational Linguistics, {ACL}}, pages 2713--2722. Association for
  Computational Linguistics.

\bibitem[{Vashishth et~al.(2018)Vashishth, Joshi, Prayaga, Bhattacharyya, and
  Talukdar}]{shikhar2018reside}
Shikhar Vashishth, Rishabh Joshi, Sai~Suman Prayaga, Chiranjib Bhattacharyya,
  and Partha~P. Talukdar. 2018.
\newblock \href {https://doi.org/10.18653/v1/d18-1157} {{RESIDE:} improving
  distantly-supervised neural relation extraction using side information}.
\newblock In \emph{Proceedings of the 2018 Conference on Empirical Methods in
  Natural Language Processing, {EMNLP}}, pages 1257--1266. Association for
  Computational Linguistics.

\bibitem[{Veli{\v{c}}kovi{\'c} et~al.(2018)Veli{\v{c}}kovi{\'c}, Cucurull,
  Casanova, Romero, Li{\`{o}}, and Bengio}]{petar2018gat}
Petar Veli{\v{c}}kovi{\'c}, Guillem Cucurull, Arantxa Casanova, Adriana Romero,
  Pietro Li{\`{o}}, and Yoshua Bengio. 2018.
\newblock \href {https://openreview.net/forum?id=rJXMpikCZ} {Graph attention
  networks}.
\newblock In \emph{Proceedings of the 6th International Conference on Learning
  Representations, {ICLR}}. OpenReview.net.

\bibitem[{Wang et~al.(2019)Wang, Focke, Sylvester, Mishra, and
  Wang}]{wang2019fine-tune}
Hong Wang, Christfried Focke, Rob Sylvester, Nilesh Mishra, and William~Yang
  Wang. 2019.
\newblock \href {http://arxiv.org/abs/1909.11898} {Fine-tune bert for docred
  with two-step process}.
\newblock \emph{CoRR}, abs/1909.11898.

\bibitem[{Wolf et~al.(2019)Wolf, Debut, Sanh, Chaumond, Delangue, Moi, Cistac,
  Rault, Louf, Funtowicz, and Brew}]{wolf2019transformers}
Thomas Wolf, Lysandre Debut, Victor Sanh, Julien Chaumond, Clement Delangue,
  Anthony Moi, Pierric Cistac, Tim Rault, R{\'{e}}mi Louf, Morgan Funtowicz,
  and Jamie Brew. 2019.
\newblock \href {http://arxiv.org/abs/1910.03771} {Huggingface's transformers:
  State-of-the-art natural language processing}.
\newblock \emph{CoRR}, abs/1910.03771.

\bibitem[{Wu et~al.(2019)Wu, Luo, Leung, Ting, and Lam}]{wu2019gda}
Ye~Wu, Ruibang Luo, Henry C.~M. Leung, Hing{-}Fung Ting, and Tak~Wah Lam. 2019.
\newblock \href {https://doi.org/10.1007/978-3-030-17083-7\_17} {{RENET:} {A}
  deep learning approach for extracting gene-disease associations from
  literature}.
\newblock In \emph{Research in Computational Molecular Biology - 23rd Annual
  International Conference, {RECOMB}}, volume 11467 of \emph{Lecture Notes in
  Computer Science}, pages 272--284. Springer.

\bibitem[{Xie et~al.(2022)Xie, Shen, Li, Mao, and Han}]{xie2022eider}
Yiqing Xie, Jiaming Shen, Sha Li, Yuning Mao, and Jiawei Han. 2022.
\newblock \href {https://doi.org/10.18653/v1/2022.findings-acl.23} {Eider:
  Empowering document-level relation extraction with efficient evidence
  extraction and inference-stage fusion}.
\newblock In \emph{Findings of the Association for Computational Linguistics:
  {ACL}}, pages 257--268. Association for Computational Linguistics.

\bibitem[{Xu et~al.(2021{\natexlab{a}})Xu, Wang, Lyu, Zhu, and
  Mao}]{xu2021entity}
Benfeng Xu, Quan Wang, Yajuan Lyu, Yong Zhu, and Zhendong Mao.
  2021{\natexlab{a}}.
\newblock \href {https://ojs.aaai.org/index.php/AAAI/article/view/17665}
  {Entity structure within and throughout: Modeling mention dependencies for
  document-level relation extraction}.
\newblock In \emph{Proceedings of the Thirty-Fifth {AAAI} Conference on
  Artificial Intelligence, {AAAI}}, pages 14149--14157. {AAAI} Press.

\bibitem[{Xu et~al.(2021{\natexlab{b}})Xu, Chen, and
  Zhao}]{xu2021discriminative}
Wang Xu, Kehai Chen, and Tiejun Zhao. 2021{\natexlab{b}}.
\newblock \href {https://doi.org/10.18653/v1/2021.findings-acl.144}
  {Discriminative reasoning for document-level relation extraction}.
\newblock In \emph{Findings of the Association for Computational Linguistics:
  {ACL/IJCNLP}}, pages 1653--1663. Association for Computational Linguistics.

\bibitem[{Xu et~al.(2021{\natexlab{c}})Xu, Chen, and Zhao}]{xu2021document}
Wang Xu, Kehai Chen, and Tiejun Zhao. 2021{\natexlab{c}}.
\newblock \href {https://ojs.aaai.org/index.php/AAAI/article/view/17667}
  {Document-level relation extraction with reconstruction}.
\newblock In \emph{Proceedings of the Thirty-Fifth {AAAI} Conference on
  Artificial Intelligence, {AAAI}}, pages 14167--14175. {AAAI} Press.

\bibitem[{Yao et~al.(2019)Yao, Ye, Li, Han, Lin, Liu, Liu, Huang, Zhou, and
  Sun}]{yao2019docred}
Yuan Yao, Deming Ye, Peng Li, Xu~Han, Yankai Lin, Zhenghao Liu, Zhiyuan Liu,
  Lixin Huang, Jie Zhou, and Maosong Sun. 2019.
\newblock \href {https://doi.org/10.18653/v1/p19-1074} {Docred: {A} large-scale
  document-level relation extraction dataset}.
\newblock In \emph{Proceedings of the 57th Conference of the Association for
  Computational Linguistics, {ACL}}, pages 764--777. Association for
  Computational Linguistics.

\bibitem[{Ye et~al.(2020)Ye, Lin, Du, Liu, Li, Sun, and Liu}]{ye2020coref}
Deming Ye, Yankai Lin, Jiaju Du, Zhenghao Liu, Peng Li, Maosong Sun, and
  Zhiyuan Liu. 2020.
\newblock \href {https://doi.org/10.18653/v1/2020.emnlp-main.582}
  {Coreferential reasoning learning for language representation}.
\newblock In \emph{Proceedings of the 2020 Conference on Empirical Methods in
  Natural Language Processing, {EMNLP}}, pages 7170--7186. Association for
  Computational Linguistics.

\bibitem[{Ye and Ling(2019)}]{ye2019distant}
Zhi{-}Xiu Ye and Zhen{-}Hua Ling. 2019.
\newblock \href {https://doi.org/10.18653/v1/n19-1288} {Distant supervision
  relation extraction with intra-bag and inter-bag attentions}.
\newblock In \emph{Proceedings of the 2019 Conference of the North American
  Chapter of the Association for Computational Linguistics: Human Language
  Technologies, {NAACL-HLT}}, pages 2810--2819. Association for Computational
  Linguistics.

\bibitem[{Yu et~al.(2022)Yu, Yang, and Tian}]{yu2022relation}
Jiaxin Yu, Deqing Yang, and Shuyu Tian. 2022.
\newblock \href {https://doi.org/10.18653/v1/2022.naacl-main.109}
  {Relation-specific attentions over entity mentions for enhanced
  document-level relation extraction}.
\newblock In \emph{Proceedings of the 2022 Conference of the North American
  Chapter of the Association for Computational Linguistics: Human Language
  Technologies, {NAACL}}, pages 1523--1529. Association for Computational
  Linguistics.

\bibitem[{Yuan et~al.(2019)Yuan, Liu, Tang, Zhang, Zhuang, Pu, Wu, and
  Ren}]{yuan2019cross}
Yujin Yuan, Liyuan Liu, Siliang Tang, Zhongfei Zhang, Yueting Zhuang, Shiliang
  Pu, Fei Wu, and Xiang Ren. 2019.
\newblock \href {https://doi.org/10.1609/aaai.v33i01.3301419} {Cross-relation
  cross-bag attention for distantly-supervised relation extraction}.
\newblock In \emph{Proceedings of The Thirty-Third {AAAI} Conference on
  Artificial Intelligence, {AAAI}}, pages 419--426. {AAAI} Press.

\bibitem[{Zaporojets et~al.(2021)Zaporojets, Deleu, Develder, and
  Demeester}]{zaporojets2021dwie}
Klim Zaporojets, Johannes Deleu, Chris Develder, and Thomas Demeester. 2021.
\newblock \href {https://doi.org/10.1016/j.ipm.2021.102563} {{DWIE:} an
  entity-centric dataset for multi-task document-level information extraction}.
\newblock \emph{Information Processing \& Management.}, 58(4):102563.

\bibitem[{Zeng et~al.(2015)Zeng, Liu, Chen, and Zhao}]{zeng2015distant}
Daojian Zeng, Kang Liu, Yubo Chen, and Jun Zhao. 2015.
\newblock \href {https://doi.org/10.18653/v1/d15-1203} {Distant supervision for
  relation extraction via piecewise convolutional neural networks}.
\newblock In \emph{Proceedings of the 2015 Conference on Empirical Methods in
  Natural Language Processing, {EMNLP}}, pages 1753--1762. The Association for
  Computational Linguistics.

\bibitem[{Zeng et~al.(2014)Zeng, Liu, Lai, Zhou, and Zhao}]{zeng2014relation}
Daojian Zeng, Kang Liu, Siwei Lai, Guangyou Zhou, and Jun Zhao. 2014.
\newblock \href {https://aclanthology.org/C14-1220/} {Relation classification
  via convolutional deep neural network}.
\newblock In \emph{Proceedings of the 25th International Conference on
  Computational Linguistics, {COLING}}, pages 2335--2344. {ACL}.

\bibitem[{Zeng et~al.(2020)Zeng, Xu, Chang, and Li}]{zeng2020double}
Shuang Zeng, Runxin Xu, Baobao Chang, and Lei Li. 2020.
\newblock \href {https://doi.org/10.18653/v1/2020.emnlp-main.127} {Double graph
  based reasoning for document-level relation extraction}.
\newblock In \emph{Proceedings of the 2020 Conference on Empirical Methods in
  Natural Language Processing, {EMNLP}}, pages 1630--1640. Association for
  Computational Linguistics.

\bibitem[{Zhang and Zhou(2014)}]{zhang2014areview}
Min{-}Ling Zhang and Zhi{-}Hua Zhou. 2014.
\newblock \href {https://doi.org/10.1109/TKDE.2013.39} {A review on multi-label
  learning algorithms}.
\newblock \emph{{IEEE} Trans. Knowl. Data Eng.}, 26(8):1819--1837.

\bibitem[{Zhang et~al.(2015)Zhang, Zheng, Hu, and
  Yang}]{zhang2015bidirectional}
Shu Zhang, Dequan Zheng, Xinchen Hu, and Ming Yang. 2015.
\newblock \href {https://aclanthology.org/Y15-1009/} {Bidirectional long
  short-term memory networks for relation classification}.
\newblock In \emph{Proceedings of the 29th Pacific Asia Conference on Language,
  Information and Computation, {PACLIC}}. {ACL}.

\bibitem[{Zhang et~al.(2018)Zhang, Qi, and Manning}]{zhang2018graph}
Yuhao Zhang, Peng Qi, and Christopher~D. Manning. 2018.
\newblock \href {https://doi.org/10.18653/v1/d18-1244} {Graph convolution over
  pruned dependency trees improves relation extraction}.
\newblock In \emph{Proceedings of the 2018 Conference on Empirical Methods in
  Natural Language Processing, {EMNLP}}, pages 2205--2215. Association for
  Computational Linguistics.

\bibitem[{Zhang et~al.(2017)Zhang, Zhong, Chen, Angeli, and
  Manning}]{zhang2017position}
Yuhao Zhang, Victor Zhong, Danqi Chen, Gabor Angeli, and Christopher~D.
  Manning. 2017.
\newblock \href {https://doi.org/10.18653/v1/d17-1004} {Position-aware
  attention and supervised data improve slot filling}.
\newblock In \emph{Proceedings of the 2017 Conference on Empirical Methods in
  Natural Language Processing, {EMNLP}}, pages 35--45. Association for
  Computational Linguistics.

\bibitem[{Zhou et~al.(2021)Zhou, Huang, Ma, and Huang}]{zhou2021document}
Wenxuan Zhou, Kevin Huang, Tengyu Ma, and Jing Huang. 2021.
\newblock \href {https://ojs.aaai.org/index.php/AAAI/article/view/17717}
  {Document-level relation extraction with adaptive thresholding and localized
  context pooling}.
\newblock In \emph{Proceedings of the Thirty-Fifth {AAAI} Conference on
  Artificial Intelligence, {AAAI}}, pages 14612--14620. {AAAI} Press.

\bibitem[{Zhu et~al.(2019)Zhu, Lin, Liu, Fu, Chua, and Sun}]{zhu2019graph}
Hao Zhu, Yankai Lin, Zhiyuan Liu, Jie Fu, Tat{-}Seng Chua, and Maosong Sun.
  2019.
\newblock \href {https://doi.org/10.18653/v1/p19-1128} {Graph neural networks
  with generated parameters for relation extraction}.
\newblock In \emph{Proceedings of the 57th Conference of the Association for
  Computational Linguistics, {ACL}}, pages 1331--1339. Association for
  Computational Linguistics.

\bibitem[{Zhu et~al.(2018)Zhu, Kwok, and Zhou}]{zhu2018multi}
Yue Zhu, James~T. Kwok, and Zhi{-}Hua Zhou. 2018.
\newblock \href {https://doi.org/10.1109/TKDE.2017.2785795} {Multi-label
  learning with global and local label correlation}.
\newblock \emph{{IEEE} Trans. Knowl. Data Eng.}, 30(6):1081--1094.

\end{thebibliography}
\bibliographystyle{acl_natbib}

\appendix

\section{Supplementary Details of Experiments}
\paragraph{Dataset} 
As stated in Section~\ref{dataset_intro}, we evaluate our model on DocRED and DWIE datasets. 
DocRED is the first large-scale crowdsourced dataset for document-level relation extraction, which is constructed from Wikipedia articles. DWIE is a dataset for document-level multi-task information extraction which contains four sub-tasks. To make it suitable for the DocRE task, the same preprocessing method as \citet{ru2021learning} is applied. 

More statistical information on training set of these two datasets is detailed in Tabel~\ref{dataset_sup}. We see that each document on DocRED contains 19.49 entities expressing 12.51 relational facts, on average. Similar observations are made on DWIE. And the number of multi-label entity pairs on two datasets are 2466 and 2880, respectively. These observations also support the multi-entity and multi-label properties of DocRE task.

\begin{table}[ht]
    \centering
    \resizebox{0.95\linewidth}{!}{
    \begin{tabular}{lcccccc}
		\toprule
		Statistics & DocRED & DWIE \\
		\midrule
		\#Relation Facts & 38180 & 14410 \\
		\#Multi-label Entity pairs & 2466 & 2880 \\
		Avg.\#Entities per Doc. & 19.49 & 27.4  \\
		Avg.\#Mentions per Entity & 1.34 & 1.97 \\
		Avg.\#Relation Facts per Doc. & 12.51 & 23.94 \\
		\bottomrule
    \end{tabular}
    }
    \caption{More statistical information of training set on DocRED and DWIE datasets.}
    \label{dataset_sup}
\end{table}

\paragraph{Parameters Setting} For the general hyper-parameters such as batch size, learning rate, etc., we set them to be consistent with \citet{zhou2021document}. For the coefficients (i.e., $\alpha$ and $\beta$) of training objectives, we search them from the list $[0.1,0.2,...,0.9]$ and pick the one with the highest F1 on the development set.

\section{Statistical Correlation Matrices}
\label{appendix_a}
\begin{figure*}[htbp]
	\centering
	\begin{minipage}{0.49\linewidth}
		\centering
		\includegraphics[width=\linewidth]{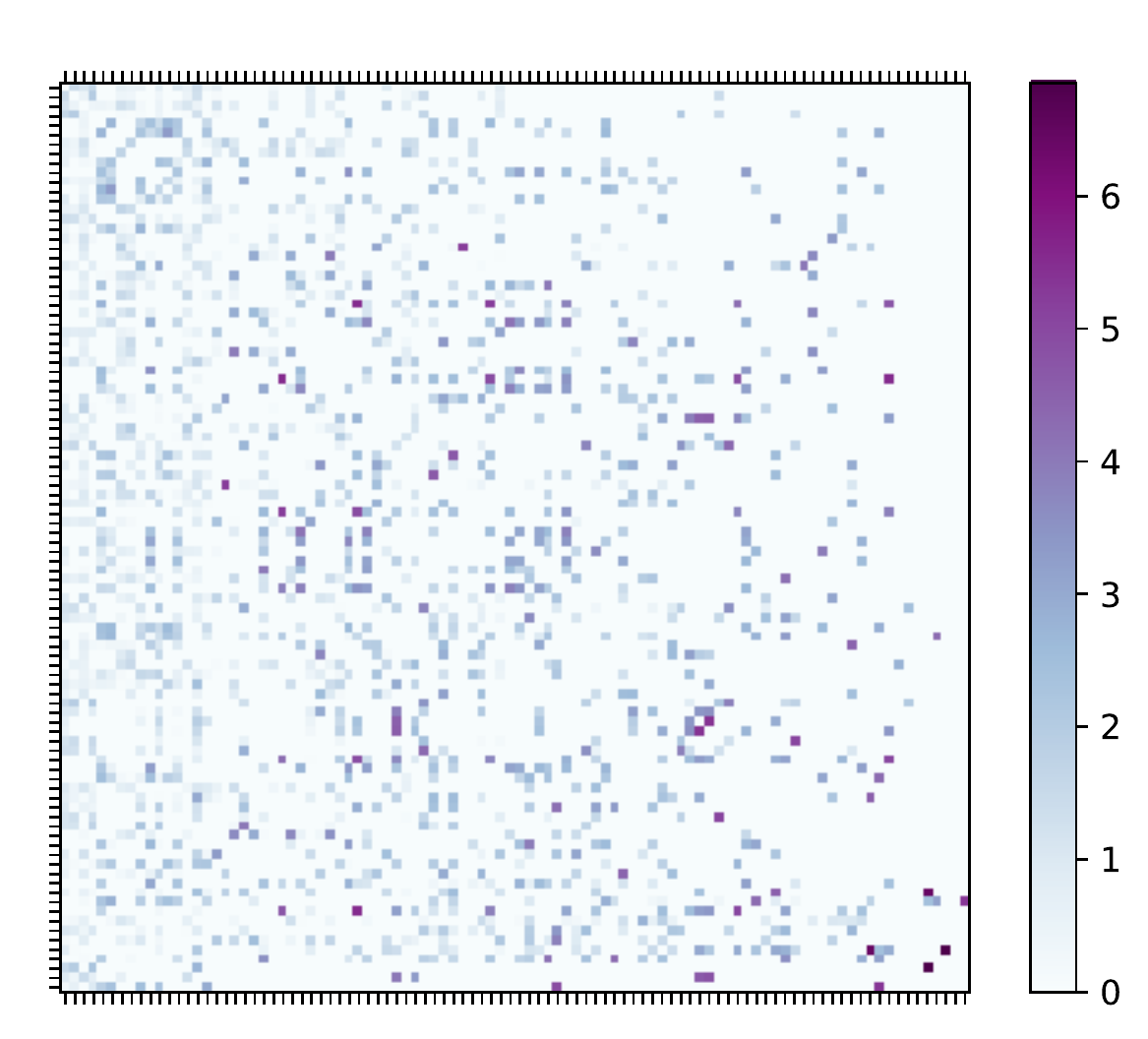}
		\label{docred_ppmi_complete}
	\end{minipage}
	\begin{minipage}{0.49\linewidth}
		\centering
		\includegraphics[width=\linewidth]{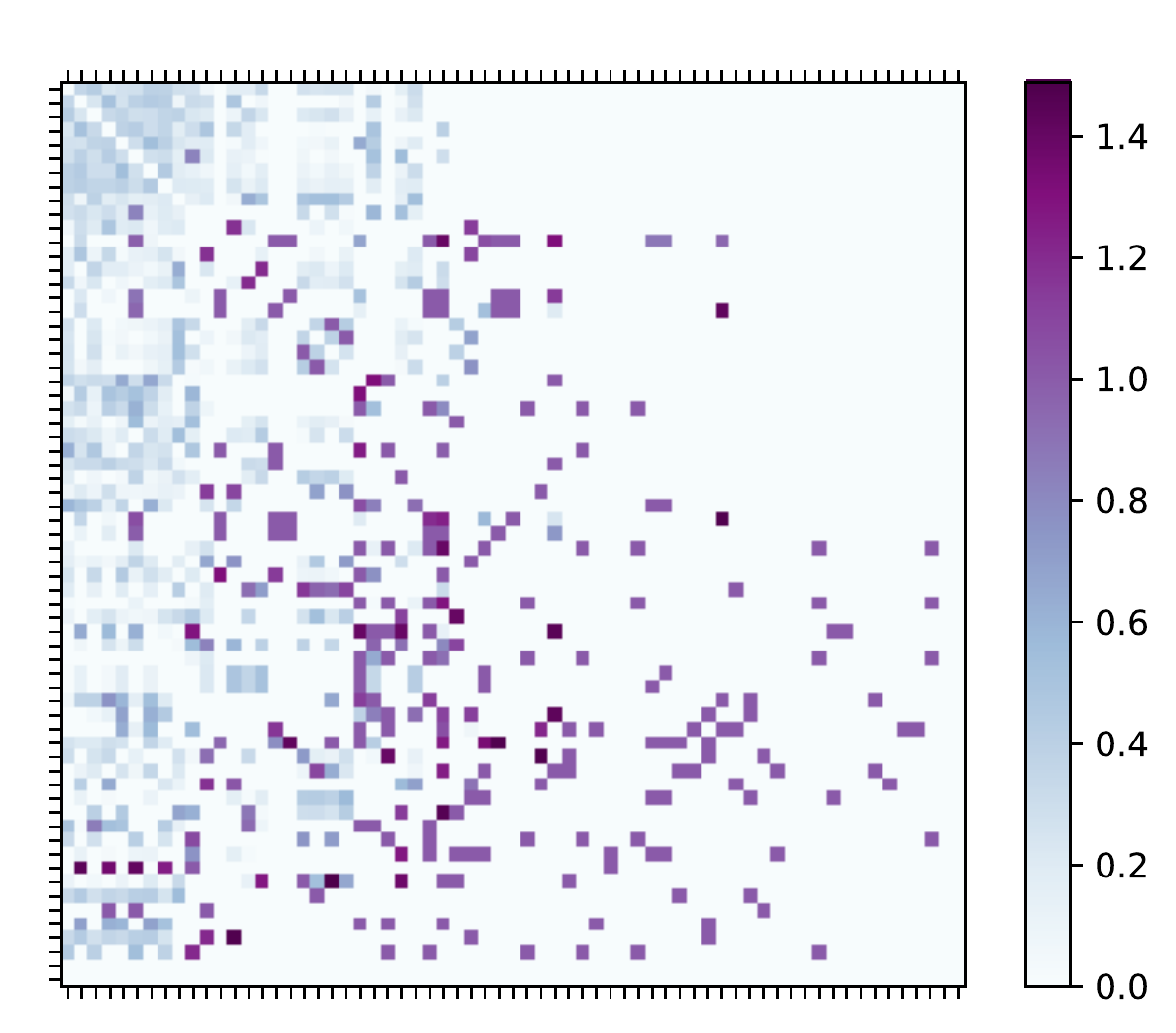}
		\label{dwie_ppmi_complete}
	\end{minipage}
    \caption{Statistical Relation correlations on DocRED (left) and DWIE (right) datasets. For better visual effect, we only display top-20 correlated relations for each row. For DWIE, we truncate all values greater than 1.5.}
    \label{ppmi_complete}
\end{figure*}

In this section, we show the complete statistical figures of relation correlations for the DocRED \cite{yao2019docred} and DWIE \cite{zaporojets2021dwie} datasets. We first count the frequency of co-occurrence between every two relations, and then calculate the positive point-wise mutual information values to measure the correlations. Relation ids are sorted in descending order by the number of training triplets.
For better visual effect, we only display top-20 correlated relations for each row. Since the correlation values of DWIE vary dramatically, we truncate all values greater than 1.5. 
In Figure~\ref{ppmi_complete}, we can find that the co-occurrence correlation of relations is widespread, which is an essential inherent characteristic of the training data itself.

\section{Long-tail Distribution of Relations}
\begin{figure*}[htbp]
	\centering
	\begin{minipage}{0.49\linewidth}
		\centering
		\includegraphics[width=\linewidth]{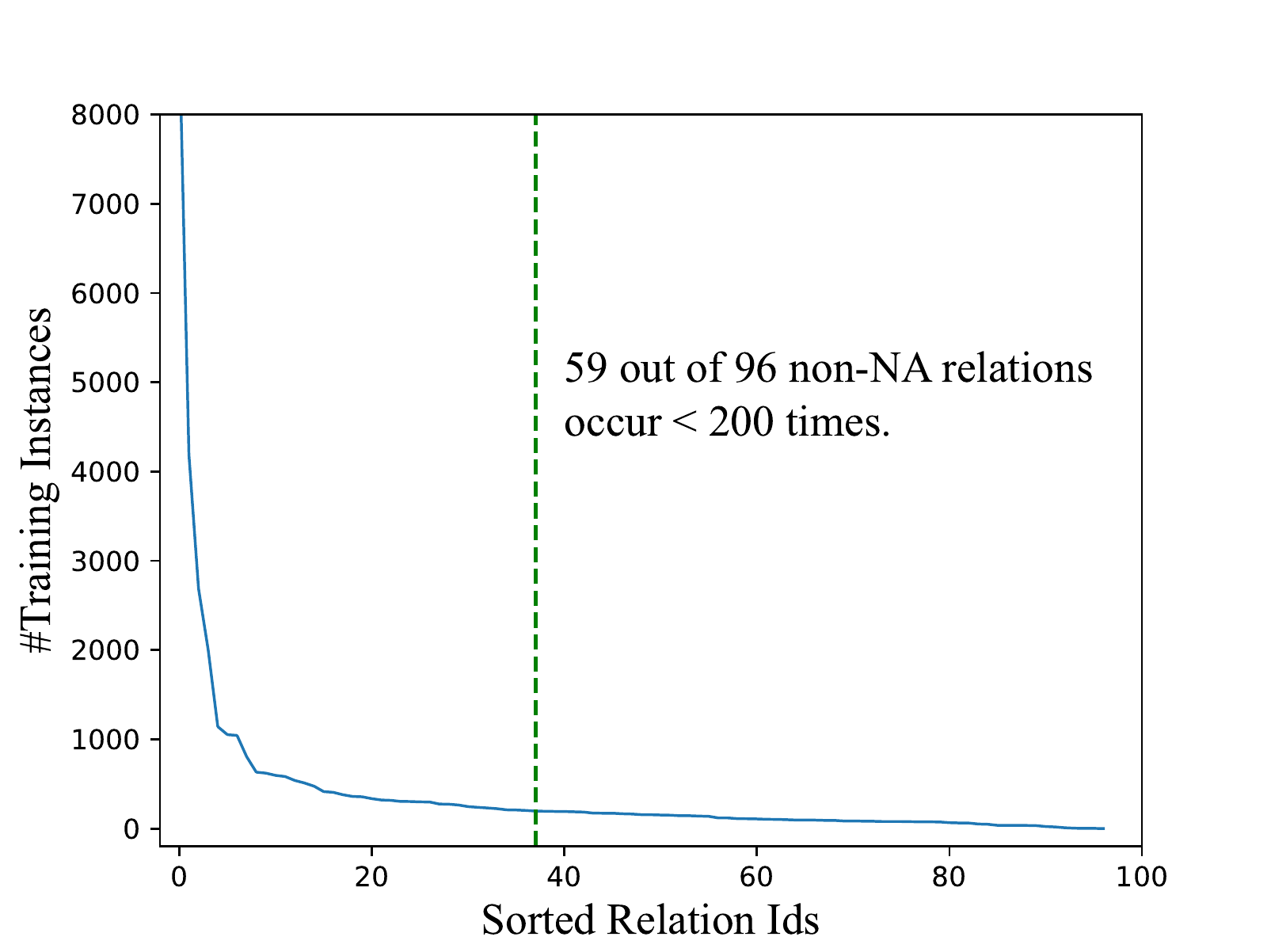}
		\label{docred_long_tail}
	\end{minipage}
	\begin{minipage}{0.49\linewidth}
		\centering
		\includegraphics[width=\linewidth]{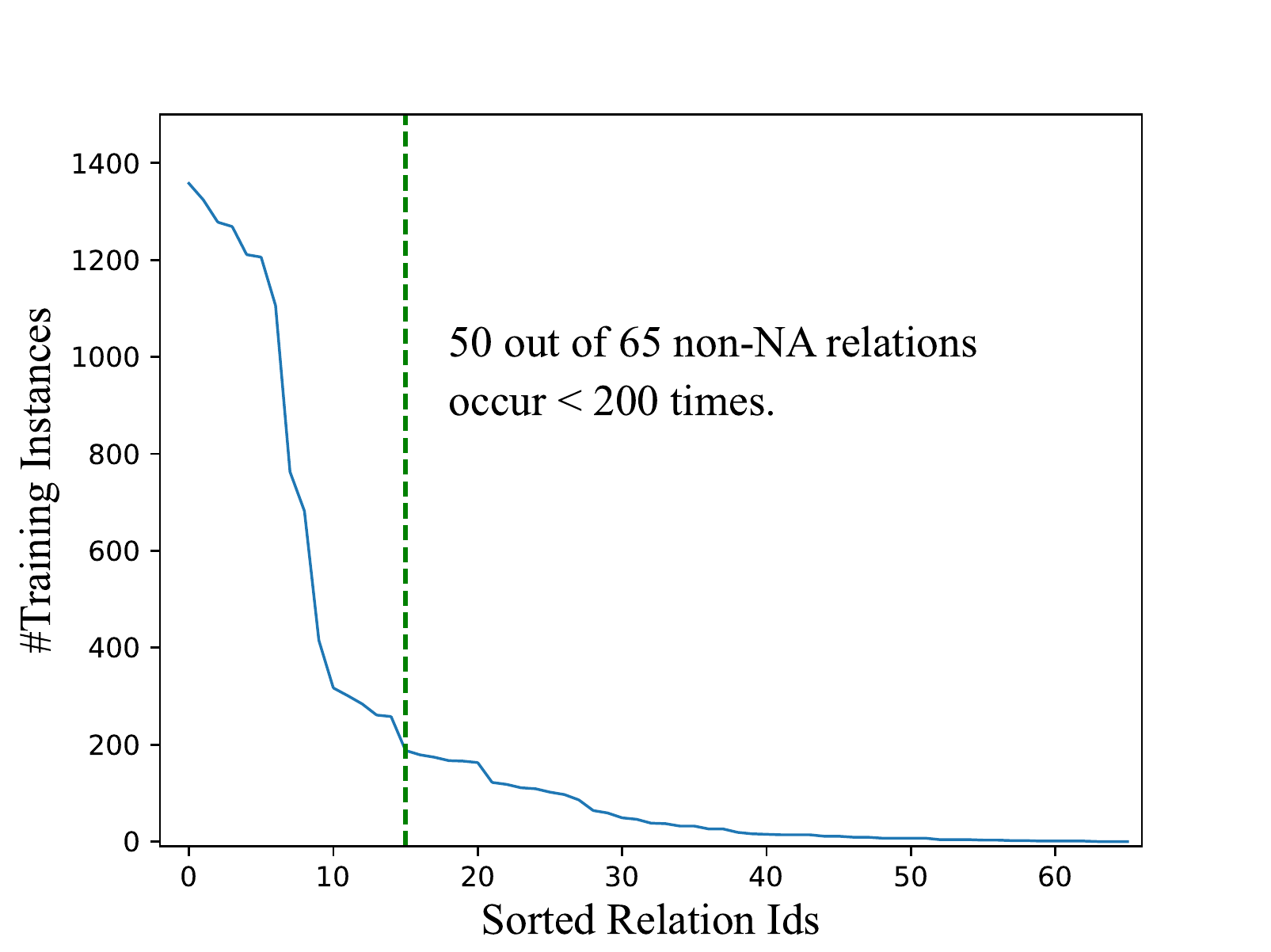}
		\label{dwie_long_tail}
	\end{minipage}
    \caption{Relation distribution of training set on DocRED (left) and DWIE (right) datasets.}
    \label{long_tail_complete}
\end{figure*}

We count the frequency of relations in the training set of DocRED and DWIE datasets, and display the distributions in Figure~\ref{long_tail_complete}. Relation ids are also sorted by frequency count from high to low. 
The green dashed lines separate data-rich relations from long-tail ones. Here the relations with less than 200 training instances are considered as long-tail relations, and vice versa for data-rich relations.
We can find that most of the relations are long-tailed on both DocRED and DWIE datasets. The lack of training instances makes these relations under-trained, severely limiting the performance of DocRE task.


\section{The Learned Correlation Matrices}
\begin{figure*}[htbp]
	\centering
	\begin{minipage}{0.49\linewidth}
		\centering
		\includegraphics[width=\linewidth]{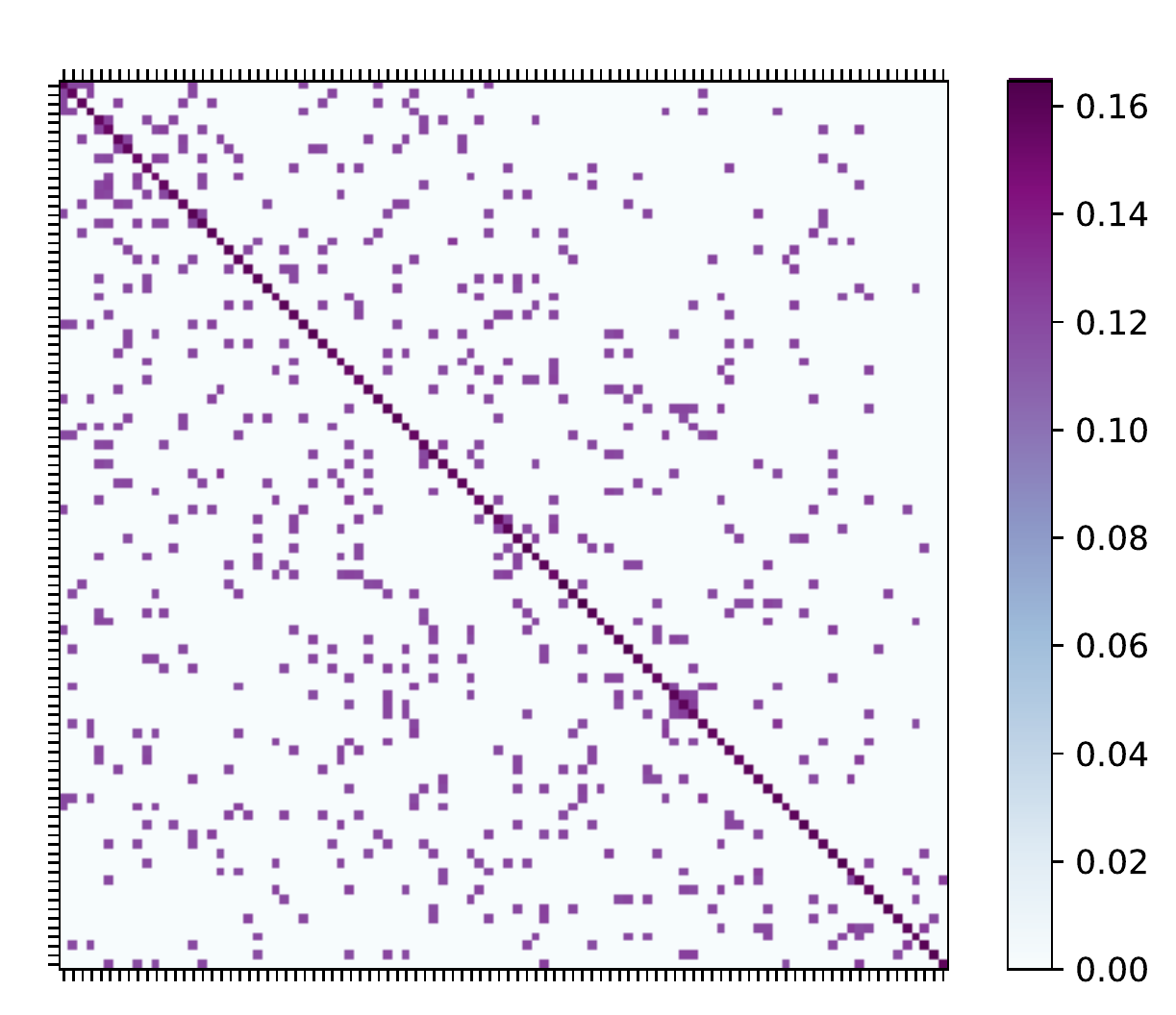}
		\label{docred_learn_complete}
	\end{minipage}
	\begin{minipage}{0.49\linewidth}
		\centering
		\includegraphics[width=\linewidth]{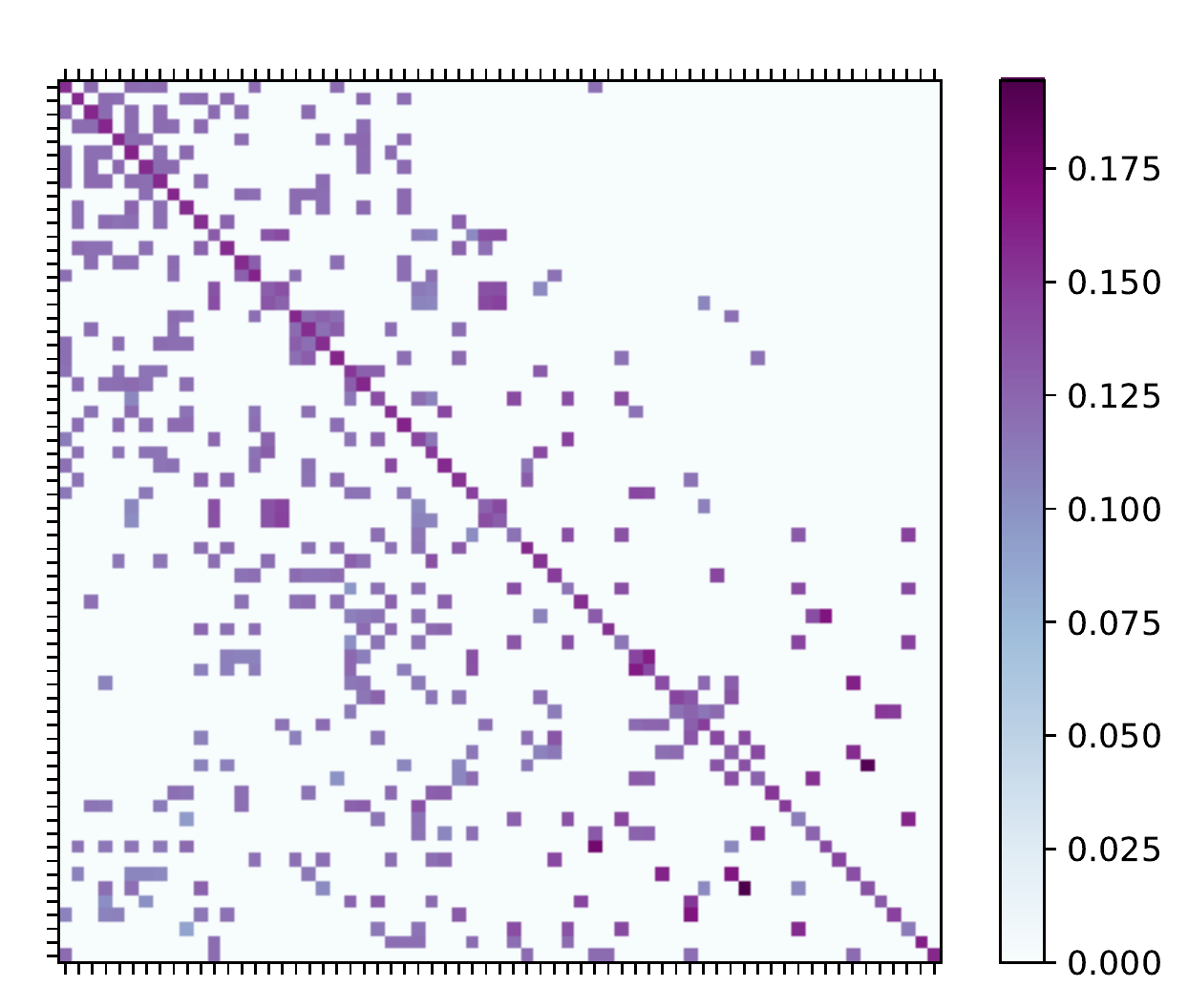}
		\label{dwie_learn_complete}
	\end{minipage}
    \caption{The learned relation correlations on DocRED (left) and DWIE (right) datasets.}
    \label{learn_complete}
\end{figure*}

To check the correlations learned by our model, we fetch the relation embeddings learned by our model on DocRED and DWIE datasets, employ dot-product to calculate the relation similarity correlation matrices, and visual the complete matrices in Figure~\ref{learn_complete}.
Same as Appendix~\ref{appendix_a}, relation ids are also sorted in descending order by the number of training triplets. 
For better visual effect, we only display top-8 correlated relations for each row. 
Although there are some noise points,
we can find that the correlation matrices we learned are very similar to Figure~\ref{ppmi_complete}. This observation implies that our model does capture large amounts of correlation knowledge.



\end{document}